\documentclass[lettersize,journal]{IEEEtran}
\usepackage{amsmath,amsfonts}
\usepackage{algorithmic}
\usepackage{algorithm}
\usepackage{array}
\usepackage[caption=false,font=footnotesize,labelfont=rm,textfont=rm]{subfig}
\usepackage{textcomp}
\usepackage{stfloats}
\usepackage{url}
\usepackage{verbatim}
\usepackage{graphicx}
\usepackage{cite}
\usepackage{multirow}
\usepackage{multicol}
\usepackage{pifont}

\begin{document}

\title{Geometric Graph Representation with Learnable Graph Structure and Adaptive AU Constraint for Micro-Expression Recognition}

\author{Jinsheng Wei,
        Wei Peng,
        Guanming Lu*,
        Yante Li,
        Jingjie Yan,
        and Guoying Zhao,~\IEEEmembership{Fellow,~IEEE}
\IEEEcompsocitemizethanks{
\IEEEcompsocthanksitem *Corresponding author: Guanming Lu (email: lugm@njupt.edu.cn).\protect
\IEEEcompsocthanksitem Jinsheng Wei, Guanming Lu and Jingjie Yan are with Nanjing University of Posts and Telecommunications (NUPT), 210003 Nanjing, China.
\IEEEcompsocthanksitem Yante Li and Guoying Zhao are with the Center for Machine Vision and Signal Analysis, University of Oulu, 90014 Oulu, Finland.
\IEEEcompsocthanksitem Wei Peng is with Department of Psychiatry and Behavioral Sciences of Stanford Medicine, Stanford University, CA 94305, USA
}

}
\maketitle
\markboth{Journal of \LaTeX\ Class Files,~Vol.~14, No.~8, August~2021}%
{Shell \MakeLowercase{\textit{et al.}}: A Sample Article Using IEEEtran.cls for IEEE Journals}


\begin{abstract}

Micro-expression recognition (MER) is valuable because micro-expressions (MEs) can reveal genuine emotions. Most works take image sequences as input and cannot effectively explore ME information because subtle ME-related motions are easily submerged in unrelated information. Instead, the facial landmark is a low-dimensional and compact modality, which achieves lower computational cost and potentially concentrates on ME-related movement features. However, the discriminability of facial landmarks for MER is unclear. Thus, this paper explores the contribution of facial landmarks and proposes a novel framework to efficiently recognize MEs. Firstly, a geometric two-stream graph network is constructed to aggregate the low-order and high-order geometric movement information from facial landmarks to obtain discriminative ME representation. Secondly, a self-learning fashion is introduced to automatically model the dynamic relationship between nodes even long-distance nodes. Furthermore, an adaptive action unit loss is proposed to reasonably build the strong correlation between landmarks, facial action units and MEs. Notably, this work provides a novel idea with much higher efficiency to promote MER, only utilizing graph-based geometric features. The experimental results demonstrate that the proposed method achieves competitive performance with a significantly reduced computational cost. Furthermore, facial landmarks significantly contribute to MER and are worth further study for high-efficient ME analysis.

\end{abstract}

\begin{IEEEkeywords}
Micro-Expression Recognition, Facial Landmarks, Graph Network, Action Units.
\end{IEEEkeywords}

\section{Introduction}%
\IEEEPARstart{F}{acial} expressions can reflect people's emotions, and representitive categories include macro-expressions and micro-expressions. Different from macro-expressions, micro-expressions (MEs) can reflect genuine emotion when people try to hide their emotions. According to the studies \cite{Pys-porter,Pys-Ekman2,Pys2-weinberger,Pys1-Yan} in psychology, MEs are human expressions with rapid, involuntary and low intensity facial movements. Considering that ME can imply the true human emotions and help reveal psychological activities \cite{Pys-porter,Pys-Ekman2}, recognizing ME gains increasing attention~\cite{XiaPeng,Wei,GACNN,Ben-2022-survey,li2022survey,li20224dme} and has valuable potential applications, \textit{e.g.} medical diagnosis \cite{Pys1-Yan}, emotion interfaces \cite{Pys-porter}, security \cite{Pys2-weinberger}, and lie detection \cite{Pys-porter,Pys-Ekman2}.

However, micro-expression recognition (MER), which aims at inferring the category of MEs from ME video clips, is very challenging due to the subtle facial muscle movements and fleeting duration. A lot of works \cite{SparseMDMO-Liu,KGSL-Zong,LGCcon-Yante,Wei-KTGSL,CEF-Peng,XiaPeng,AU-ICGAN} extracted discriminative features based on RGB images/videos through machine learning models, \textit{e.g.}, LBP-TOP with SVM \cite{LBPTOP-Zhao,LBP-FIP} or deep neural networks~\cite{LGCcon-Yante}. Though promising results are achieved, there is still a big space to improve in terms of both accuracy and efficiency. Due to the suppressed property of MEs, only small parts of facial region will be activated in a ME instance. Thus, huge efforts are wasted to deal with redundant information of a whole image/video. Even worse, the ME-related features are prone to be submerged in irrelevant information, thereby making MER more difficult. To solve this problem, this work proposes to extracts discriminative geometric features only from facial landmarks. In fact, when facial muscle moves, facial landmarks will move accordingly. Thus, facial muscle movements in ME can be captured by geometric features in facial landmarks \cite{Survey2018-Yee}. At the same time, we argue that the extremely compact geometric information is good enough to get comparable even may superior performance, compared to the redundant feature information based on image sequences.
Furthermore, time and computational costs regarding landmark-based methods are greatly less than that of image-based methods, which is more desirable for practical applications like in mobile devices. Currently, few works \cite{LFM,G-TCN,GACNN} considered facial landmark information to improve the performance of image-based appearance features, getting promising performance. However, these works did not focus on studying the contribution of geometric features from facial landmarks to MER. Thus, in this paper, we make the first step to study the discriminability of facial landmarks for MER and employ only landmarks as input, for exploring its advantages compared to image-based input.

The graph-based methods\cite{liu2021graphsurvey} have been verified to process the facial landmarks lying in the  non-Euclidean space. Lately, some existing works \cite{G-TCN,Lei2,GACNN,MER-auGCN} employed the graph-based models to recognize MEs. However, these works adopt a fixed graph structure designed manually according to certain principles (\textit{e.g.} the conditional probability of different nodes \cite{Lei2,MER-GCN}). Namely, the fixed adjacency matrix is defined to represent the fixed relationship between nodes. The manual way to set a fixed graph structure cannot comprehensively and effectively model the correlation between landmark-based nodes, which is sub-optimal. For overcoming this problem, this paper introduces a learnable adjacency matrix to learn a more reasonable and flexible graph structure based on a pre-defined graph structure.

Action Unit (AU) \cite{FACS} encodes the local movement of facial muscles and have a strong correlation with MEs. Recently, some works \cite{AU-ICGAN,MER-auGCN} took AU features as input to recognize MEs and used deep learning models to build the relationship between AU features and ME categories. However, these methods need an auxiliary model to extract AU features before inputting AU features into the main model, which increases the model complexity and computational cost. To overcome this defect, our work uses the AU labels as a constraint to learn the geometric features related to AUs from facial landmarks.
Furthermore, considering the AU losses in different layers may have different contribution, this paper proposes an adaptive AU loss to automatically learn the constraint intensity of AU loss in different layers. In this way, the multi-scale geometric features are constrained adaptively to have a synchronized pattern with AUs, thereby rationalizing the introduction of facial action units (AUs) information.


Corresponding to the above problems, the main contributions of this paper can be summarized as follows:
\begin{itemize}
  \item [1)]
This work studies and explores the contribution of facial landmarks for MER. The graph-based models are employed to extract discriminative geometric movement features with spatial-temporal information only from facial landmarks. The results demonstrate that facial landmarks are discriminative for MER with a largely reduced computational cost. 
  \item [2)]
To comprehensively explore geometric movement features from facial landmark, a Geometric Two-Streams Graph Network (GTS-GN) is proposed to aggregate the low-order and higher-order geometric information from facial landmarks.
  \item [3)]
To overcome the shortcoming of the fixed adjacency matrix, this paper proposes a Learnable Adjacency Matrix (LAM) to learn a reasonable and flexible graph structure. As a result, LAM can automatically model the discriminative relationship between facial landmarks for MER.

  \item [4)]

Based on the strong correlations between  AUs and MEs, an Adaptive AU (AAU) loss is proposed to automatically explore a more reasonable and efficient way to introduce AU information. AAU loss can adaptively constrain the geometric features in a multi-scale fashion and emphasize the contributions of AU information at different semantic levels. The experimental results demonstrate that jointing the regular ME loss and AAU loss can build the relationship between facial landmarks, AUs and MEs, improving the performance of MER. 

\end{itemize}

\section{Related Work}

With the research of MER, increasing number of researchers have focused on automatic ME analysis, expecting to explore its value and potential in both academic research and commercial applications. Based on traditional machine learning methods, early works \cite{SMIC-Xiaobai,MDMO-Liu,DLBP-RIP-Huang,HOG-TOP-Polik} classified MEs using hand-crafted features. These hand-crafted features require effort to design, and lacks adaptability. Later, deep learning methods were well applied in image processing fields, and then some works \cite{GEME,DSSN-Khor} in MER began to explore deep learning methods and achieved promising performance. In addition to classical deep learning models, \textit{e.g.} Convolutional Neural Network (CNN), some recent works \cite{G-TCN,GACNN,MER-auGCN,Lei2,MER-GCN,AU-ICGAN} tried to use graph-based models to recognize MEs. The experimental results have proved the effectiveness of the graph-based methods in MER.

\subsection{Input Datas}
So far, many works \cite{SMIC-Xiaobai,DLBP-RIP-Huang,Wei,DSSN-Khor} extract discriminative features from the whole facial RGB images/videos or the feature maps based on these images/videos. However, the extracted features include a lot of redundant information because the discriminative features only exist in a small area of the face. The classification model spends too much energy in areas that are not related to MEs, and requires too much computational cost. Zong \textit{et al.}\cite{KGSL-Zong} proposed the kernelized group sparse learning (KGSL) model to select effective regions, improving the recognition rate. However, it still requires extracting several hierarchical features from the whole face image, thereby increasing computational cost. To overcome the above problem, Polikovsky \textit{et al.} \cite{HOG-TOP-Polik} divided the Regions of Interest (ROI)s from the whole facial region based on the facial landmarks, and then the features from these ROIs are input into classifier to get ME categories. Some works \cite{HOG-TOP-Polik,Hybrid-Region-Liong} verified the effectiveness of the features extracted only from ROIs with a reduced computational cost. Furthermore, Liu \textit{et al.} \cite{MDMO-Liu} proposed the main direction of optical flow (MDMO) to represent the main direction of movement in ROIs. The more compact features are input to SVM classifier. Utilizing local regions, instead of the global ones, these works largely reduce the costs. However, these local region-based methods still take up computing resources to extract the features in the local area. For some discriminative features, the computational cost still is huge, \textit{e.g.} Optical Flow. In addition, region cropping is also a not trivial issue. 

As a comparison, facial landmarks are low-dimension data, and taking it as the input can reduce computational costs. In MER, facial landmarks need to be extracted for face alignment, interception, and the division of regions, especially the division of local regions in local region-based methods. Thus, extracting facial landmarks is unavoidable for MER. At the same time, when facial muscle moves, facial landmarks will move accordingly. Facial muscle movements related to MEs can be captured by the geometric features in facial landmarks. The movwment of facial landmarks can represent the movement information of facial muscle in key regions. Therefore, we argue that geometric information is discriminative enough to get comparable even superior performance. Facial landmarks-based methods may not be necessary to perform complicated and costly pre-processing from the raw RGB inputs. Compared with images/videos, the low dimensional landmarks are much more compact, as well can include the discriminative geometric feature information\cite{OER3}. Choi \textit{et al.} \cite{LFM} employed facial landmarks as the input to recognize MEs instead of original ME images/videos. Specifically, they proposed 2D landmark feature map (LFM) by transforming conventional coordinate-based landmark information into 2D image information. Then, CNN and Long Short Term Memory (LSTM) models are employed to process LFM. The work aggregated geometric feature information in facial landmarks and got promising results. However, the model input still is the 2D images. 
Recently, Kumar \textit{et al.}\cite{GACNN} proposed a two-stream Graph Convolutional Network (GCN)\cite{GCN} model to process the optical flow and facial landmark information. The above works introduce the facial landmark information into the model. However, these works do not focus on landmarks or explore their discriminability and efficiency. Thus, this paper focuses on the effectiveness of facial landmarks in terms of both accuracy and efficiency. Furthermore, we explore more effective modules and components to aggregate spatial and temporal information in facial landmarks. Since the landmarks are structure data lying in the Non-Euclidean space, a new graph model is designed to learn the geometric feature representations of MEs from the facial landmarks.

\subsection{Graph Models}
Recently, graph-based methods are popular in MER and generally don't input the whole facial images into models. According to the way to construct graph, these methods can be divided into two types: landmark-based and AU-based.

Lei \textit{et al.} \cite{G-TCN} firstly constructed a landmarks-based graph. The magnified shape features around landmarks were taken as node features. However, they ignore the special relationships of nodes and only employed Temporal Convolutional Network (TCN) \cite{TCN} to deal with the node representation. Recently, Kumar \textit{et al.} \cite{GACNN} took the facial landmarks and optical flow features around landmarks as node features, and designed Graph Attention Convolutional Network (GACN) to process the landmark-based graph. Except for facial landmarks, AU also is suitable to construct the graph. Lo \textit{et al.} \cite{MER-GCN} designed GCN with two layers to deal with AU-based graph and aggregate AU label information that dots product with spatial-temporal features extracted by 3D convolution. They explores the relationship between AUs and MEs, but the node features are the annotated AU labels, which is not suitable for practical applications. Furthermore, both \cite{AU-ICGAN} and \cite{MER-auGCN} took the AU features extracted by extra model as node features in AU-based graph. Differently, Zhou \textit{et al.} \cite{AU-ICGAN} connected the inception block and an AU detection module to extract AU features, while Xie \textit{et al.} \cite{MER-auGCN} employed 3D ConvNet to extract AU features. Lei \textit{et al.} \cite{Lei2} introduce transformer and AU features to extend their previous work \cite{G-TCN} through combining two types of graph.
However, on one hand, the existing works with landmark-based graph aggregated dynamic information, \textit{e.g.} optical flow and magnified shape feature, which still spends too much computational cost to extract these features. On the other hand, the works with AU-based graph need extra learning model to extract AU features as node feature, which increases the model complexity and computational cost. Also, AU detection for MER itself is a very challenging task. Inspired from skeleton-based action recognition works, \textit{e.g.} \cite{ST-GCN-AR}, instead of providing much complicated and costly appearance features or not easily accessible AU features, we provide a simple and much efficient way that directly takes landmark coordinate-based geometric features as node features. A new graph network is designed to deal with such geometric features from facial landmarks for MER.
\begin{figure*}
\setlength{\belowcaptionskip}{0pt}
\setlength{\abovecaptionskip}{0pt}
\begin{center}
\includegraphics[width=1\linewidth]{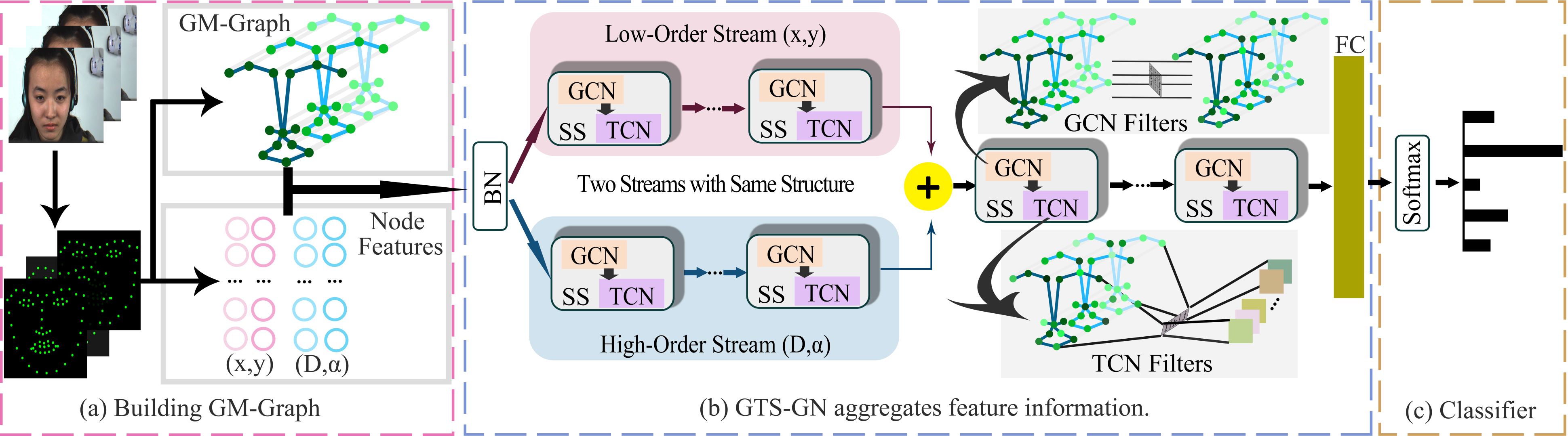}
\end{center}
   \caption{The flowchart of GTS-GN. GTS-GN aggregates the geometric information of low order coordinate $(x,y)$ and high order semantic ($D$, $\alpha$).
   }
\label{TS-GCN}
\vspace{-3mm}
\end{figure*}
Furthermore, the proposed graph-based method involves the relationship between nodes and the AU information. The related works are as follows:

\subsubsection*{The Relationship between Nodes}
The relationship between nodes is key to aggregate the feature information in graph. For a graph, the adjacency matrix can express the node relationship. Although Lei \textit{et al.} \cite{G-TCN} defined a graph, they didn't give the adjacency matrix in the graph. Namely, the special relationship between nodes do not be considered. Most of the above graph-based methods built the relationship between nodes according to some principles. Specifically, based on the human facial structure, Kumar \textit{et al.} \cite{GACNN} built the relationship between landmark nodes. \cite{MER-auGCN} first connected AU nodes based on the AU correlations of the objective class definition and then adjusted it referring to the training datasets. \cite{AU-ICGAN} achieved it by the training set based on common facial expressions and facial anatomy analysis. \cite{Lei2} and \cite{MER-GCN} defined the adjacency matrix by a data-driven way, and both use the conditional probability of different nodes. Different from these methods, we first pre-defined a fixed adjacency matrix based on facial muscle structure. Then, we define a LAM that is added to the fixed adjacency matrix. LAM can be updated automatically to learn a more reasonable relationship between nodes.

\subsubsection*{AU Information}
As we know, AUs have the closed relation with facial expressions, including MEs. At present, AU recognition received widespread attention, and the researchers in MER have introduced AU information to recognize MEs. In \cite{MER-GCN} and \cite{Lei2}, AUs are fed into the GCN model to learn the AU representations. The learned AU representations are fused with ME representations extracted by other models. However, the two methods need to compute the AU vector in advance. \cite{MER-auGCN} and \cite{AU-ICGAN} introduced AU loss to extracted AU representation from original video data or optical flow map. Then, the learned representations are aggregated by GCN for final classification. These methods need the extra models to process AUs or learn AU representation, which increases the model complexity and computational cost. Different from these works, the proposed AAU loss aggregate AU information without extra models. It can constrain multi-scale features in an adaptive way to reasonably construct the strong relationship between facial landmarks, AUs and MEs.

\section{The Proposed Method}

In this section, the proposed facial landmark-based method is introduced in detail. We present the framework of the proposed graph model as shown in Fig. \ref{TS-GCN}. Specifically, this work constructs a geometric movement graph (GM-Graph), designs SS module to deal with GM-Graph, and builds the GTS-GN model. In addition, two key components (LAM and AAU loss) are proposed to be applied to the designed module and model. 
The details are as follows.

\subsection{Geometric Movement Graph}

\begin{figure}
\setlength{\belowcaptionskip}{0pt}
\setlength{\abovecaptionskip}{6pt}
\centering
\subfloat[Spatial graph $G_o$ for a single frame.]{\includegraphics[height=3.5cm,width=8cm]{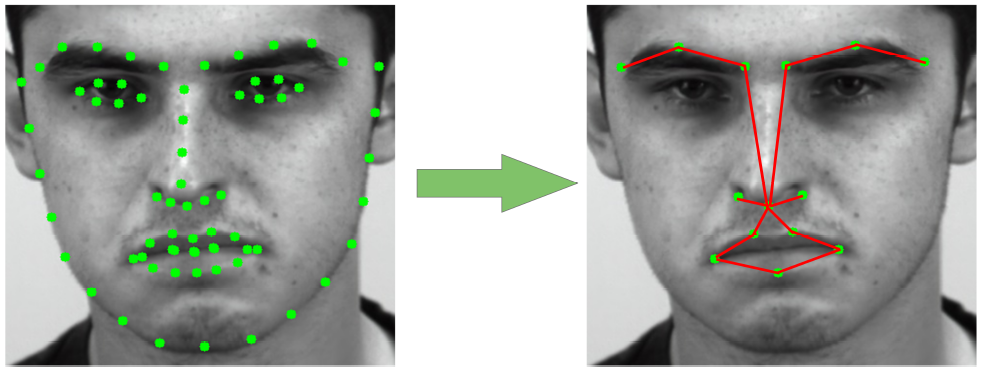}
\label{GO}}
\vspace{-1mm}

\subfloat[GM-Graph]{
\includegraphics[height=3.5cm,width=4.7cm]{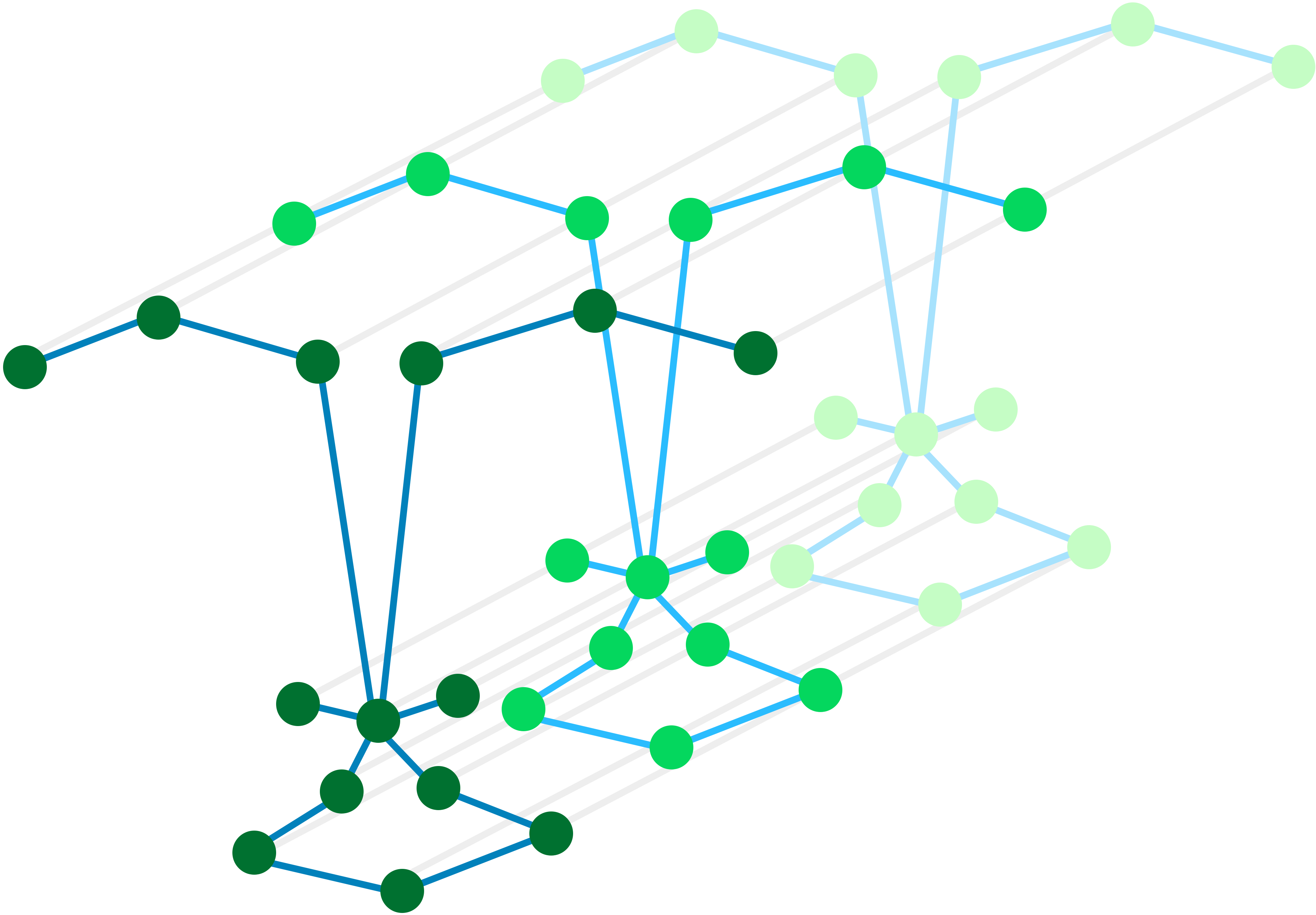}
\label{ST-G}
}
\subfloat[$D$ and $\alpha$]{
\label{SN}
\includegraphics[height=3.5cm,width=3.15cm]{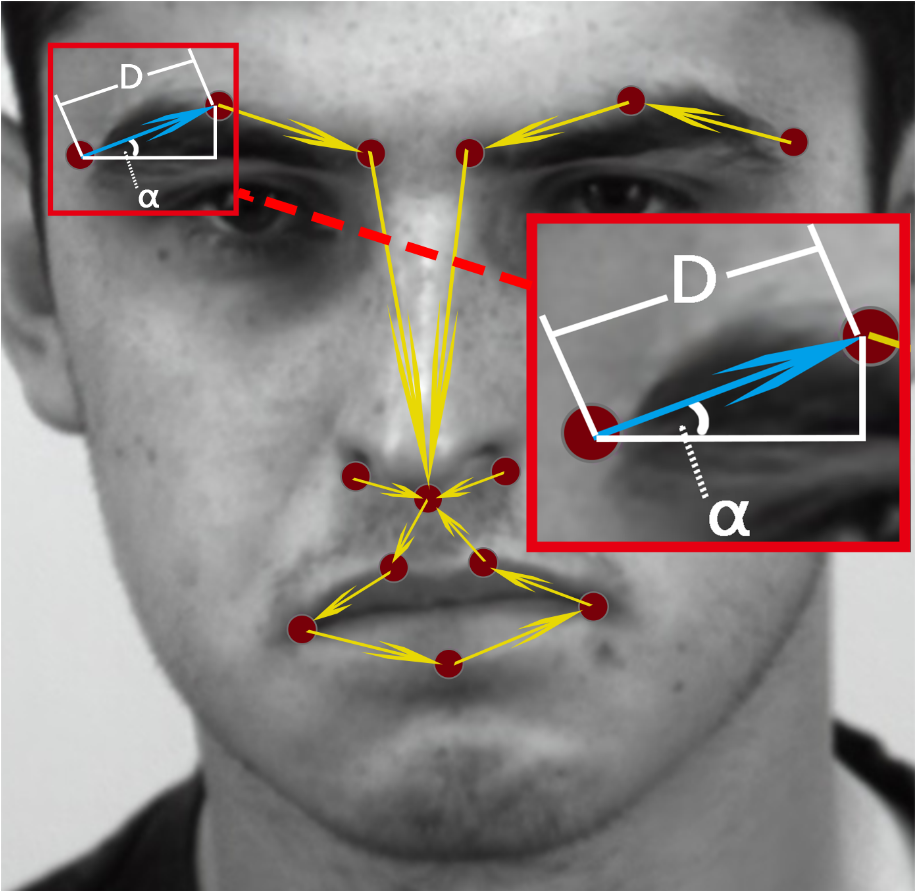}
}
\caption{The construction of GM-Graph based on onset, apex, and offset frames, and the calculation of the distance $D$ and angle $\alpha$. 
}
\label{CCF}
\vspace{-5mm}
\end{figure}

In fact, the extraction of movement features is crucial to MER tasks. For a ME video, the onset, apex and offset frames can represent the key process of muscle movement related to MEs. Thus, these three frames contain abundant movement information and remove a lot of redundant information in the entire ME video.

Accordingly, the facial landmarks of the onset, apex and offset frames include the discriminative geometric movement information. Thus, in this paper, the landmarks of the onset, apex and offset frames in ME videos are taken as nodes in GM-Graph to capture the ME movements as shown in Fig. \ref{CCF}. Furthermore, inspired by the Facial Action Coding System (FACS)~\cite{FACS}, we choose the landmarks around the mouth, eyebrows and nose to construct the graph based on their contributions. The landmarks in the eye region don't be considered here. The reasons are as follows: 1) some noises in the eye regions exist, \textit{e.g.} blinking. 2) the change in the eye regions about MEs can lead to the corresponding change in eyebrows, \textit{e.g.} dilation of pupils in a surprised expression. So, the movements in the eye region have a certain amount of redundant information.

As shown in Fig. \ref{GO}, we select 14 key points as the nodes of the spatial graph from facial landmarks. Considering the natural connection of the facial regions, we build the spatial relationship between landmarks for a single frame. Therefore, a landmark-based spatial graph $G_o=\{N_o,E_o\}$ is constructed, where $N_o$ and $E_o$ are the node set and edge set in spatial graph, respectively. In fact, movement features need to be extracted from temporal information that is key to recognize MEs. Thus, based on the $G_o$ of the onset, apex and offset frames, the GM-Graph $G_{GM}$ is constructed to establish the spatial-temporal relationship between facial landmarks. As shown in Fig. \ref{ST-G}, for modeling temporal information, the corresponding nodes between three frames also are connected. Thus, GM-Graph $G_{GM}=\{N,E\}$ is constructed, where $N$ and $E$ are the node set and edge set in GM-Graph, respectively, $N$ is a collection of the $N_o$ of three frames, and $E$ denotes the connection of nodes in GM-Graph.

The node features are crucial to represent the spatial-temporal information. In fact, the movements of facial muscles will cause the corresponding movement of facial landmarks. Furthermore, the captured ME videos need a high quality due to the subtle facial movement, which ensures high accuracy of detecting facial landmarks. Thus, the landmark coordinates include the key movement information in a ME instance. The data dimension of landmark coordinates is much smaller than that of the raw frames, which is more efficient. This advantage of landmark coordinates can save computational resources to facilitate practical applications. In this paper, only landmark coordinates $n=(x,y)$ are adopted as nodes features to study the effectiveness of landmarks, and higher-order semantic features (distance and angle between landmarks) are added to explore the interaction of low and high-order geometric information.

Fig. \ref{SN} shows the calculation of distance $D$ and angle $\alpha$ based on landmark coordinates. Suppose that $n_i=(x_i,y_i)$ (the start point of the arrow in Fig. \ref{SN}) is the current node, and $n_j=(x_j,y_j)$ (the end point of the arrow) is the neighboring node. $D$ and $\alpha$ can be calculated by:
\begin{equation}
\begin{cases}
D=(x_i-x_j)^2+(y_i-y_j) ^2\\
\alpha=arctan((y_i-y_j)/(x_i-x_j))
\end{cases},
\end{equation}
where $arctan(*)$ is the arctangent function.

According to the description above, we define two types of node features that $(x,y)$ is Type A, and $(x,y,D,\alpha)$ is Type B, where $(x,y)$ is the landmark coordinates.

\begin{figure}
\setlength{\belowcaptionskip}{0pt}
\setlength{\abovecaptionskip}{0pt}
\begin{center}
\includegraphics[width=1\linewidth]{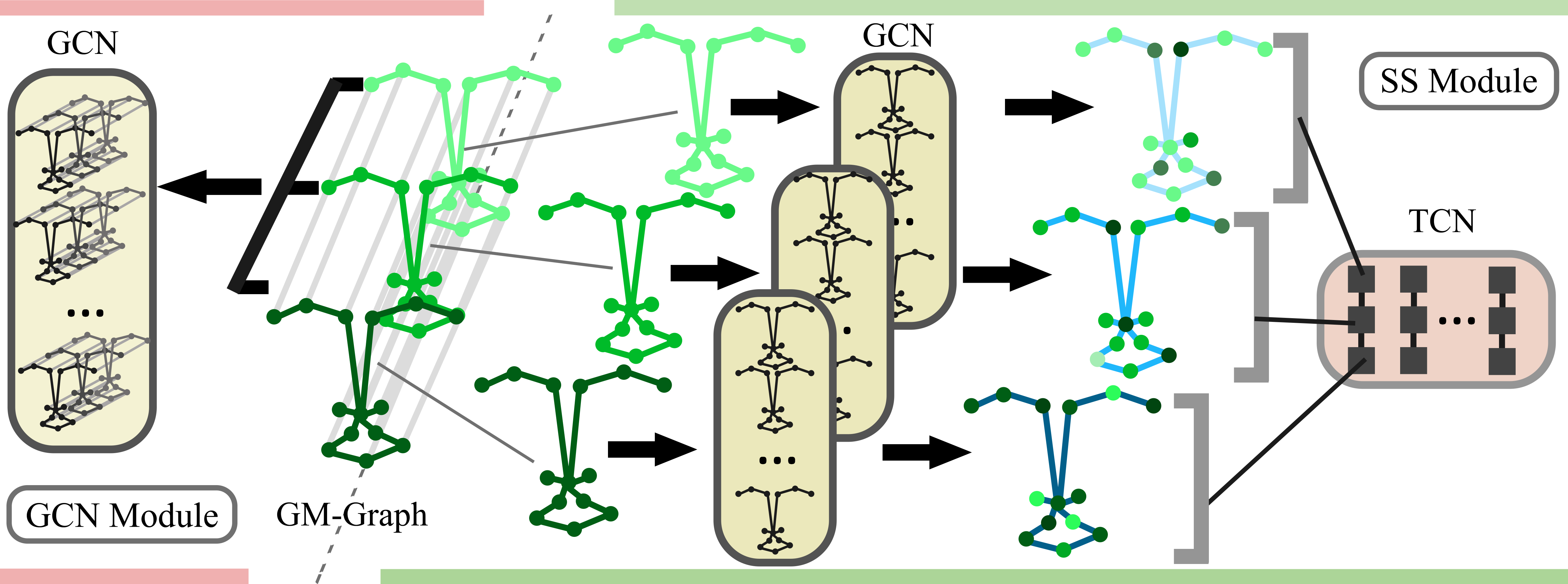}
\end{center}
   \caption{The comparison of GCN and SS Modules.}
\label{M}
\vspace{-5mm}
\end{figure}

\subsection{SS Module}

Inspired by 
CNN+LSTM~\cite{LFM} that extracts spatial-temporal features separately, we design SS module to aggregate spatial and temporal information in GM-Graph. The movement of the landmarks in ME is small, and the movement of each node along the time step is key to recognize MEs. Thus, the extraction of temporal features is crucial and challenge for MER. GCN can simultaneously extract the spatial and temporal features in the spatial-temporal graph. However, it cannot focus on the extraction for the temporal features, which maybe neglect some small movement features. Thus, in this work, TCN is adopted to focus on temporal information in GM-Graph. Furthermore, since the spatial features of each frame also includes important information, \textit{e.g.} geometric and structure information, GCN is employed to aggregate spatial information in GM-Graph. As shown in Fig. \ref{M}, SS module adopts GCN to aggregate the spatial information for three frames, respectively. Then, TCN is adopted to aggregate the temporal information between three frames. Like that, every operation focuses on the aggregation of one type of information (spatial or temporal), which facilitates more finely learning geometric movement features.

\subsubsection*{GCN with the Learnable Adjacency Matrix}
Specifically, SS module first splits GM-Graph into three sub-graphs \{${G_{GM^1},G_{GM^2},G_{GM^3}}$\}. The three sub-graphs correspond to the spatial graph $G_o$ of three frames. Suppose that $A$ is the pre-defined adjacency matrix that expresses the node connection in $G_o$, and $A \in \mathbb{R}^{14*14}$. By normalizing $A$, we can get $L=I_n-D^{-1/2} A D^{-1/2}$, where $D=\sum_{j} A_{ij}$ is the degree matrix.
Then, $L$ is used for Fourier transform. For $G_{GM^f} (f=1,2,3)$ with spatial information, the graph is filtered by $g_\phi$ to get the output node features $Y^f$ as follows:
\begin{equation}
\begin{split}
Y^f=g_\phi(L)X=Ug_\phi(\Lambda)U^TX^f, 
\end{split}
\end{equation}
where $X^f$ is the node features of $G_{GM^f}$; $U$ is the Fourier basis and a set of orthonormal eigenvectors for $L$; and $L=U \Lambda U^T$ with $\Lambda$ as corresponding eigenvalues. As the calculation of eigenvectors matrix is expensive, a Chebyshev polynomial with $R$-th order is employed to well-approximate the filter $g_\phi$ as follows:
\begin{equation}
\begin{split}
Y^f=\sum_{r=1}^{R} {\theta_r}C_r(\hat{L})X^f,
\end{split}
\end{equation}
where $\theta_r$ denotes Chebyshev coefficients, and $C_r(\hat{L})$ denotes Chebyshev polynomial. $\hat{L}=2L/\lambda_{max} -I_n$ is normalized to [-1,1], and $C_r(\hat{L})=2\hat{L}C_{r-1}(\hat{L})-C_{r-2}(\hat{L})$, where $C_0 = 1$ and $C_1 = \hat{L}$. Suggested by \cite{GCN}, $R=1$, and $\lambda_{max}=2$. Then,
\begin{equation}
\begin{split}
Y^f={\theta_0}X^f - \theta_1 (D^{-1/2} A D^{-1/2})X^f \\
= \theta(I_n+D^{-1/2} A D^{-1/2})X^f,
\end{split}
\end{equation}
where $\theta_r$ is approximated to a unified $\theta$, and $\theta=\theta_0=-\theta_1$. Finally, for simplifying expression, we set $L=I_n+D^{-1/2} A D^{-1/2}$ and the final form of GCN is:
\begin{equation}
\begin{split}
Y^f=GCN(G_{GM^f})=LX^f\theta,
\end{split}
\label{GCN}
\end{equation}
where $\theta$ is the learnable filter. 

\textbf{Learnable Adjacency Matrix LAM}: The above pre-defined adjacency matrix $A$ is fixed and expresses a fixed relationship between nodes. Furthermore, the fixed relationship is defined according to some principles (\textit{e.g.} facial structure \cite{GACNN} and data-driven \cite{Lei2,MER-GCN}). These principles are set by the researchers and are sub-optimal. Thus, we introduce LAM expressed as $A_L$ to learn a more reasonable relationship between nodes. Considering that the fixed $A$ has certain rationality, $A$ is retained. Therefore, the final form of GCN with LAM is as follows:
\begin{equation}
\begin{split}
Y^f=GCN^{\prime}(G_{GM^f})=(L+A_L)X^f\theta,
\end{split}
\label{GCN}
\end{equation}
where $A_L$ can be updated and learned automatically in the training stage of model.

\subsubsection*{TCN}
After getting the spatial graph representation processed by GCN with LAM, TCN is employed to extract temporal features between \{$G_{GM^1}, G_{GM^2}, G_{GM^3}$\}. 
So, the temporal feature $F^T_n$ of $n$-th node is:
\begin{equation}
\begin{split}
F^T_n=TCN(Y^1_n,Y^2_n,Y^3_n)=Y_n W,
\end{split}
\label{GCN}
\end{equation}
 where $Y_n=[Y^1_n,Y^2_n,Y^3_n]$, and $W$ is the learnable filter. Thus, SS module can be formed by:
\begin{equation}
\begin{split}
TCN(GCN^{\prime}(G_{GM^1}),GCN^{\prime}(G_{GM^2}),GCN^{\prime}(G_{GM^3})).
\end{split}
\end{equation}

As basic blocks, several SS modules are stacked directly to build SS Graph Network (SS-GN). Also, based on SS modules, GTS-GN is built as follows.

\begin{figure*}
\setlength{\belowcaptionskip}{0pt}
\setlength{\abovecaptionskip}{0pt}
\begin{center}
\includegraphics[width=0.78\linewidth]{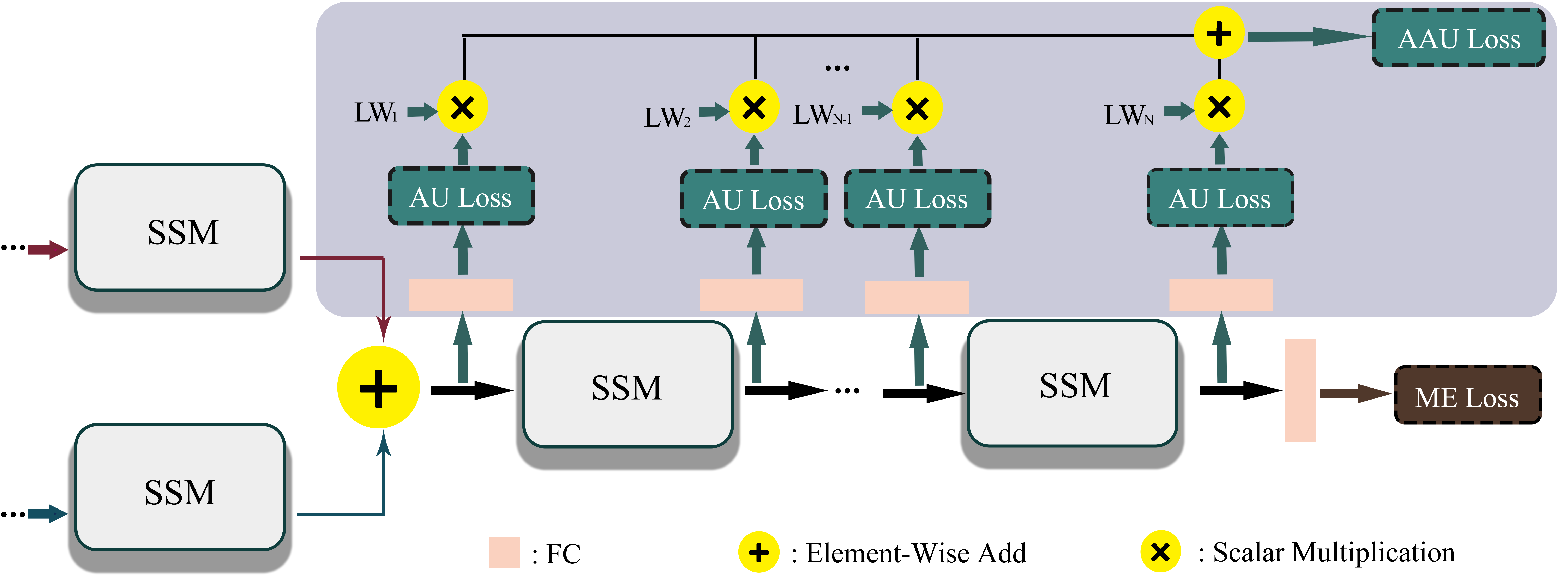}
\end{center}
   \caption{The calculation of the total loss. In this figure, LWr and N are $\frac{W_r^2 L_r}{\sum_{r=1}^{N_L} W_r^2}$  and $N_L$, respectively. }
\label{LOSS}
\vspace{-5mm}
\end{figure*}

\subsection{Geometric Two-Stream Graph Network}

Besides the low-order coordinates, the geometric features also include the high-order semantic features, \textit{e.g.} distance $D$ and angle $\alpha$. The distance and angle of landmarks include semantic information. It can provide more discriminative features to improve performance. However, it is difficult to learn the discriminative features from the information interaction of the low-order and high-order representations since the two features have different distributions. Training two models to deal with two features separately cannot take into account the interaction of the two features. Also, two models trained independently is not end-to-end. Thus, an end-to-end model that is able to aggregate low-order and high-order features separately is needed. To meet it, this paper proposes a novel graph model called GTS-GN to process low-order and high-order geometric features in two streams. 

Both features processed by GTS-GN belong to geometric features, and there is a certain correlation between the two features. Thus, an earlier fusion of the two features may be more appropriate for information interaction. However, the existing two-stream models fuse two features in last layer, \textit{e.g.} \cite{GACNN}. Different from these works, GTS-GN tries to fuse the two features at the earlier layer, not limited to the last layer. The whole network is shown in Fig. \ref{TS-GCN}.

Specifically, the low-order coordinates and high-order semantic features have some differences in feature distribution, so we first employ Batch Normalization (BN) to normalize the two features. Next, two types of features are inputted into two streams. Two steams adopt the same structure that stacks several SS modules with the same number. After several SS modules, the outputs of two streams are added together. The added features are inputted into several SS modules or Full Connected (FC) layer to continue aggregate geometric feature information. Finally, the softmax is used to classify features and predict the ME categories.

\subsection{Adaptive AU Loss}
Recent researches \cite{AU-ICGAN,MER-auGCN} have verified that the correlation between AUs and MEs is strong. Using the strong correlation is beneficial to recognize MEs \cite{MER-auGCN}. Furthermore, the geometric movement information in facial landmarks closely relates to the local movement represented by AUs. Thus, our model builds this correlation by learning the geometric movement features related to AUs before the ME classification layer. In this paper, the loss functions are employed to achieve this intention by constraining feature way. 

For AU recognition, multi-label AU loss can constrain the learned features in the classification layer so that they are relevant to AUs for classifying AU categories. As opposed to recognizing AUs, our aim is to constrain the learned features before the ME classification layer so that they are relevant to AUs for classifying ME categories. Thus, we introduce multi-label AU loss before the ME classification layer and ME loss in the classification layer. On one hand, before ME classification layer, using multi-label AU loss can map the geometric features of facial landmarks to AU features. On the other hand, in the ME classification layer using ME loss function can map AU features to high-level ME features.

Furthermore, the multi-scale features in multiple layers are constrained by the proposed AAU loss as shown in Fig. \ref{LOSS}. In order to better aggregate low-order and high-order information individually before fusion, AAU loss constrains the geometric features after fusing the two streams. Now, supposed that the features of $N_L$ layers are constrained, and the corresponding multi-label AU loss is $L_r (r=2,...,N_L)$. So, the loss to constrain the multi-scale features is as follows:
\begin{equation}
\begin{split}
L^{\prime}=\sum_{r=1}^{N_L} L_r,
\end{split}
\label{origLoss}
\end{equation}
where $L_r$ is conducted by $MultiLabelSoftMarginLoss$ package in Pytorch \cite{Pytorch}.
However, in Equation \ref{origLoss}, the constraint strengths in all layers are the same, which can't consider the difference between the features of different scales. Thus, we introduce the learnable weights to these losses for emphasizing the contributions of the features at different semantic levels, as follows:
\begin{equation}
\begin{split}
L^{\prime\prime}=\sum_{r=1}^{N_L} W_r L_r,
\end{split}
\label{Lossv1}
\end{equation}
 where $W_r$ is learnable weight and can be updated in training stage.
In fact, Equation \ref{Lossv1} make the model tend to learn the $W_r$s with all zero values. Furthermore, the positive weights also is needed. Thus, to overcome the two problems, we take probability form as the weight to get AAU Loss, as follows:
\begin{equation}
\begin{split}
L_{AAU}=\sum_{r=1}^{N_L} \frac{W_r^2 L_r}{\sum_{r=1}^{N_L} W_r^2},
\end{split}
\label{Lossv2}
\end{equation}
where $N_L\ge2$ due to AAU loss will degenerates to AU loss when $N_L=1$.


\textbf{Total Loss}: The task for this work is MER. Thus, ME labels are employed to calculate cross-entropy loss as ME loss ($L_{ME}$) in the final classification layer. Thus, the total loss ($L_T$) is as follows:
\begin{equation}
\begin{split}
L_T=L_{ME}+\beta*L_{AAU},
\end{split}
\label{Lossv3}
\end{equation}
where $\beta$ is the trade-off parameter.

\section{Experiments}

This section reports the experimental results. We study the effectiveness of landmarks and evaluate the performance of the proposed method. First, the effectiveness of facial landmarks is studied for MER. Second, we carry out ablation analysis with some visualizations to evaluate the proposed model and components, including SS module, GTS-GN, LAM, and AAU loss. Third, the parameter evaluation is carried out; Finally, we compare the proposed method with state-of-the-art (SOTA) methods.

\subsection{Experimental Setting}

\begin{table}
\setlength{\belowcaptionskip}{0pt}
\setlength{\abovecaptionskip}{0pt}
\caption{The categories of the used samples.}
\centering
\footnotesize
\setlength{\tabcolsep}{2pt}
\subfloat[\footnotesize On CAMSE \uppercase\expandafter{\romannumeral2}]{
\begin{tabular}{|l|p{20pt}|c|c|c|c|c|c|}
\hline
\multicolumn{8}{|c|}{Based on AUs}\\
\hline
Type &I  & II &III &IV &V &VI  &VII\\
\hline
\ding{172}&\checkmark&\checkmark&\checkmark&\checkmark&\checkmark&&\checkmark\\
\ding{173}&\checkmark&\checkmark&\checkmark&\checkmark&\checkmark& & \\

\hline\hline
\multicolumn{8}{|c|}{Based on self-report}\\
\hline
Type &happy  & Disgust &Repression &Surprise &Sad &Fear  &Others\\
\hline
\ding{174}&\checkmark&\checkmark&\checkmark&\checkmark&& & \checkmark\\
\hline

\end{tabular}
       \label{AA:1}
}

\vspace{-2mm}
\subfloat  [\footnotesize On SAMM ]{
\begin{tabular}{|l|c|c|c|c|c|c|c|c|}
\cline{1-8}
\multicolumn{8}{|c|}{Based on AUs}\\
\cline{1-8}
Type &I  & II &III &IV &V &VI  &VII\\
\cline{1-8}
\ding{172}&\checkmark&\checkmark&\checkmark&&&&\checkmark\\
\ding{173}&\checkmark&\checkmark&\checkmark&\checkmark&\checkmark& & \\
\cline{1-8}
\hline
\multicolumn{9}{|c|}{Based on self-report}\\
\hline
Type &happy  & Disgust &anger &Surprise &Sad &Fear& contempt &Others\\
\hline
\ding{174}&\checkmark&&\checkmark&\checkmark&&&  \checkmark& \checkmark\\
\hline
\end{tabular}
\label{}
}
\label{Dataset}
\vspace{-3mm}
\end{table}

\begin{table}
\setlength{\abovecaptionskip}{0pt}
\setlength{\belowcaptionskip}{0pt}
\vspace{0pt}
\caption{The feature dimension of every layer in all graph models used in our experiment. c is the number of categories.}
\begin{center}
\footnotesize
\begin{tabular}{|l|c|c|c|c|c|}
\hline
layer & 1&2 &3&4&FC \\
\hline
 Input dimension  & 2 or 4 & 64&64&128 &128 \\
Output dimension  & 64& 64  &128&128& c\\
\hline
\end{tabular}
\end{center}
\label{SM}
\vspace{-5mm}
\end{table}

The experiments are carried out on SAMM \cite{SAMM-Davison} and CASME \uppercase\expandafter{\romannumeral2} \cite{CASME2-Yan} datasets. In CASME \uppercase\expandafter{\romannumeral2}, all participants are from one ethnicity, and it contains 255 video samples from 26 subjects with seven ME categories based on self-report.
In SAMM, participants are from 13 ethnicities, and it includes 159 samples from 32 subjects with eight ME categories based on self-report.
The video samples in both datasets are collected by the high-speed cameras at 200 fps. Both datasets contain AU annotations that can form seven ME categories based on AU annotations;

In this work, the ME categories based on AU annotations are adopted to evaluate the proposed method. Like in previous MER tasks~\cite{LGCcon-Yante,Wei}, due to the long-tail distribution of the samples, we deleted the categories with less than ten samples. As a result, six and four ME categories based on AU annotations (Type \ding{172} in Table \ref{Dataset}) are selected from CASME \uppercase\expandafter{\romannumeral2} and SAMM, respectively. Furthermore, the existing methods used not only AU annotations, but also self-reported annotations. Thus, for a more comprehensive and fair comparison with SOTA methods, we report the results of the proposed method on other two types of category setting: under five ME categories based on AU annotations (Type \ding{173} in Table \ref{Dataset}) following \cite{ObClass-TL-Peng} and under five ME categories based on self-report (Type \ding{174} in Table \ref{Dataset}) following \cite{GACNN}. 

As in the existing works, \textit{e.g.} \cite{liu2019neural,G-TCN}, the onset, apex and offset frames can be found based on labels on databases, and their detection belongs to another task \cite{Surveyspottingkh,SpotA1,LGCcon-Yante} in ME analysis.
Furthermore, the facial muscle movements are magnified by a learning-based motion magnification method~\cite{LVMM}, such that the movements of landmarks are more obvious. The amplification factor is set to three following the setup in \cite{G-TCN}. In addition, the Dlib \cite{Dlib} package is employed to detect facial landmarks. 
To study the effectiveness of landmarks, the basic graph-based models with landmark are compared to the basic CNN model with image: ResNet18\cite{ResNet-He}. All graph-based models used in our experiment have four layers, and the feature dimensions in different layers are shown in Table \ref{SM}. $\beta$ is set to [0.1,1] with interval 0.1 in parameter evaluation and is fixed to 1 in the ablation study. To explore a better way to fuse the features from the two-stream inputs, GTS-GN is implemented in four types according to in which layer the two streams are merged, namely GTS-GN (layer $n$), $n=1,2,3,4$. The filter size of TCN is three to cover the node features in three frames. Two evaluation matrices under leave-one-subject-out (LOSO) strategy are adopted to evaluate different methods, including accuracy(ACC) and F1-score.

Data augmentation is important for the training of the deep models.
In our experiments, ResNet18 take the magnified apex frame or three key frames as inputs. For ResNet-based models, data augment is performed by rotation, adding noise, cropping randomly, horizontal flip and color jitter. 
Also, the detected landmarks for the face regions with different sizes have a small difference. Thus, we regard this difference as noise to augment data. The aligned frames are cropped to get the frames of different sizes. Then, we detect the landmarks of these frames with different sizes to obtain more data. All models are trained on a single GTX 1050 GPU with Pytorch \cite{Pytorch} deep learning framework.

\begin{table}

\setlength{\abovecaptionskip}{0pt}
\setlength{\belowcaptionskip}{0pt}
\caption{Comparison of ResNet18 (S), ResNet18 (ST), GCN and SS-GN. ACC(\%).}
\centering
\footnotesize
\setlength{\tabcolsep}{10pt}
\subfloat[\footnotesize The performance comparison.]{
\begin{tabular}{|l|c|c|c|c|}
\hline
\multirow{2}{*}{Methods} &\multicolumn{2}{c|}{CAMSE \uppercase\expandafter{\romannumeral2}} &\multicolumn{2}{c|}{SAMM} \\
\cline{2-5}
&ACC &F1 &ACC &F1\\
\hline\hline
ResNet18 (S) & 66.13  & 0.588& 68.57& 0.649\\
ResNet18 (ST) &71.31& 0.669& 72.13& 0.686\\
GCN\cite{GCN} &70.12& 0.690 & 70.21 & 0.661\\
SS-GN &\textbf{72.90}& \textbf{0.716} & \textbf{75.17} & \textbf{0.732}\\
\hline
\end{tabular}
\label{LD:P}
}
\vspace{-1mm}

\subfloat[\footnotesize The costs. Average Time (Unit: sec.)]{
\begin{tabular}{|l|c|c|c|c|}
\hline
Methods &Input size &Average Time&Parameter \\
\hline\hline
ResNet18 (ST) & 3*224*224 &0.0309& 11,179,590 \\
GCN\cite{GCN} & 42*2& 0.0038 &39,760\\
SS-GN &14*3*2& 0.0050& 163,710\\
\hline
\end{tabular}
\label{LD:ET}
}
\label{LD}
\vspace{-5mm}
\end{table}

\subsection{The Study on the Effectiveness of Landmarks}

First of all, by comparing with image, we evaluate the effectiveness of landmarks in terms of recognition performance and model efficiency.
For fair comparison, we use vanilla models to process images and landmarks. In terms of image-based methods, basic CNN models are adopted to process video frames that is Euclidean data. Correspondingly, basic graph models are adopted to process facial landmarks that is Non-Euclidean data. Specifically, image-based ResNet18 is compared with landmark-based GCN and SS-GCN. Image-based methods include ResNet18 (S) and ResNet18 (ST). ResNet18 (S) and ResNet18 (ST) adopt the same model (ResNet18), with different inputs. Specifically, for ResNet18 (S), the apex frame is the three-channel image. For ResNet18 (ST), onset, apex, and offset frames are first converted to three grayscale images, respectively. Then, the three grayscale images are taken as three channels of the input, respectively. Thus, ResNet18 (S) can aggregate spatial information, while ResNet18 (ST) can aggregate spatial and temporal information.

Table \ref{LD}(a) reports the performance comarison between ResNet18 (S), ResNet18 (ST), GCN and SS-GN. Compared with the other three methods, ResNet18 (S) has the poorest performance, which demonstrates that temporal information has an important contribution to improve MER performance. Compared with image-based ResNet18 (S) and ResNet18 (ST), landmark-based SS-GCN gets substantially better performance under both evaluation metrics. For instance, SS-GCN improves the ACC by 1.59\% and 3.04\%, compared with ResNet18 (ST), on CASME \uppercase\expandafter{\romannumeral2} and SAMM, respectively. It demonstrates that the magnified movement of facial landmarks can effectively represent the muscle movement related to MEs. In addition, ME images include redundancy information, which may decrease the MER performance. Instead, facial landmark is a more compact modality which can retain discriminative geometric features for MER and achieves promising performance. 

For the model efficiency, as shown in Table \ref{LD}(b), SS-GCN can be about 6.18 times faster than image-based ResNet18, respectively. In particular, the model size of SS-GCN is about only 1.5\% of that of ResNet18 (ST), which is essential to real-time applications. In addition, GCN and SS-GCN have a lower input dimension compared with ResNet18 (ST). These results demonstrate that compared with image-based methods, landmark-based methods have an obvious advantage in calculation cost and efficiency.

To summarize, the results above demonstrate that the movements of facial landmarks contain the discriminative movement information for MER. The landmark-based methods can effectively recognize MEs. The below experiment (the results comparing with other methods are as shown in Tables \ref{O5} and \ref{E5}) can also demonstrate the discriminability of facial landmarks for MER tasks. Also, the geometric features in facial landmarks are more compact representations with a largely reduced computational cost. Overall, compared with the image-based methods, the landmark-based graph methods have much higher computational and parameter efficiency with a competitive recognition rate.

\begin{table}
\setlength{\abovecaptionskip}{0pt}
\setlength{\belowcaptionskip}{0pt}
\caption{The evaluation on SS module and GTS-GN. ACC(\%).}
\centering
\footnotesize
\setlength{\tabcolsep}{3pt}
\subfloat[\footnotesize Graph models with different node features.]{
\begin{tabular}{|l|p{20pt}|c|p{20pt}|c|}
\hline
\multirow{2}{*}{Methods} &\multicolumn{2}{c|}{CAMSE \uppercase\expandafter{\romannumeral2}} &\multicolumn{2}{c|}{SAMM} \\
\cline{2-5}
&ACC  &F1 &ACC &F1\\
\hline\hline
GCN+Type A & 70.12& 0.690 & 70.21 & 0.661\\
SS-GN+Type A &72.91& 0.716 & 75.17 & 0.732\\
SS-GN+Type B & 72.91  & 0.710& 73.05& 0.693\\
GTS-GN+Type B & \textbf{73.31}& \textbf{0.717}& \textbf{75.89}&\textbf{0.741}\\

\hline
\end{tabular}
       \label{SSM}}
\vspace{-1mm}

\subfloat[\footnotesize GTS-GN (n layer), n = 1, 2, 3, 4.]{
\begin{tabular}{|l|p{20pt}|c|c|p{20pt}|c|}
\hline
\multirow{2}{*}{Methods} &\multicolumn{2}{c|}{CAMSE \uppercase\expandafter{\romannumeral2}} &\multicolumn{2}{c|}{SAMM} \\
\cline{2-5}
&ACC  &F1 &ACC  &F1\\
\hline\hline
GTS-GN (layer 1) & \textbf{73.31}& \textbf{0.717}& \textbf{76.60}& 0.729\\
GTS-GN (layer 2) & 72.91& 0.714& 75.17& 0.723\\
GTS-GN (layer 3) & 71.71& 0.690& 73.05& 0.708\\
GTS-GN (layer 4) & 68.13& 0.640& 75.89& \textbf{0.741}\\
\hline
\end{tabular}
\label{GTS-GN}
}
\label{AS-Model}
\vspace{-5mm}
\end{table}

\noindent\textbf{Discussions}: Notably, our method is based on landmarks. The accuracy of detecting landmarks is key for our method. In fact, if the detected landmarks have a huge error, the methods that need to use landmarks will be significantly affected. On one hand, in MER task, all existing movement-based methods need to detect landmarks to align faces. The crop of local areas in the local feature-based methods\cite{TCN, Lei2, GACNN, MDMO-Liu} also depends on landmarks. Thus, this problem is inevitable in MER task. On the other hand, for the used datasets, the scenarios are controlled and rather simple, so landmarks can be reliably detected. But when moving to uncontrolled environment with complicated mixed movement and big illumination changes, more powerful landmark detection technology is needed for getting accurate landmarks. Fortunately, the detection technologies of landmarks are ongoing research, and some works \cite{LandMarkDT2} have promising results, which is technical support for landmark-based methods. In addition, image-based methods also are greatly affected in complex environments, thus, MER under complex environments is another unsolved and challenging task.

Overall, the landmark-based methods face the above-mentioned inevitable problems like other image-based methods. Although image-based methods can aggregate more appearance information, the landmark-based methods can avoid the cumulative error caused by cropping the ROI areas and extracting features in these areas. In addition, this paper demonstrates the discriminability and efficiency of landmarks. Yet the geometric features from landmarks and the appearance features from ME images do not conflict. Therefore, how to better combine the appearance features with the geometric features is also worthy of further study.

\subsection{Ablation Analysis}
This section reports and analyzes the results of ablation study for the proposed components. It can be divided into three sub-sections: the evaluation on SS-module and GTS-GN, the evaluation on LAM, and the evaluation on AAU Loss.

\subsubsection{The Evaluation on SS-module and GTS-GN}

\noindent\textbf{The Evaluation on SS-module}: To evaluate SS module, the vanilla GCN \cite{GCN} is taken as baseline and directly processes the whole GM-Graph. As shown in Table \ref{AS-Model}(a), taking landmark coordinates (Type A) as node features, SS-GN outperforms GCN in terms of both accuracy and F1-score for both datasets. It demonstrates that the features extracted by SS module are more discriminative than those extracted by GCN. Also, extracting spatial and temporal geometric features separately is a better choice for MER.  

\noindent\textbf{The Evaluation on GTS-GN}: We evaluate the performance of GTS-GN in terms of aggregating low-order and high-order geometric features. Type B (x,y,D,$\alpha$) is taken as node features. According to Table \ref{AS-Model}(a), SS-GN+Type A is superior to SS-GN+Type B, which proves that simply introducing high-order geometric information cannot ensure performance improvement. GTS-GN+Type B outperforms both SS-GN+Type B and SS-GN+Type A. It demonstrates that GTS-GN can aggregate the low-order and high-order geometric information more effectively than SS-GN. Also, distance $D$ and angle $\alpha$ provide the discriminative high-order geometric information to recognize MEs. Furthermore, we explore a better way to fuse two feature flows in GTS-GN. As listed in Table \ref{AS-Model}(b), interestingly, we find that different datasets prefer different ways to fuse the features. For CASME \uppercase\expandafter{\romannumeral2}, fusing the two-stream features at layer one provides better performance, while for SAMM, fusing at fourth layer provides a better F1-score and a comparable accuracy. This may be caused by the different ME categories in these two datasets. The experiment results demonstrate that the fusing two feature flows in the last layer is not optimal, and earlier fusion may result in better performance.

Overall, SS module can effectively aggregate spatial-temporal information in GM-Graph, and the GM-Graph with only geometric features as node features can be processed by graph model to get promising results. GTS-GN provides flexibility to fuse and aggregate effectively the low-order and high-order geometric features. Additionally, two stream-based GTS-GN is more effective than one stream-based SS-GN. 

\subsubsection{The Evaluation on Learnable Adjacency Matrix}

\begin{figure*}
\vspace{-5mm}
\setlength{\abovecaptionskip}{0pt}
\setlength{\belowcaptionskip}{0pt}
\begin{center}
\footnotesize
\subfloat [\footnotesize Layer 1]{
\label{LAM1}
\includegraphics[width=0.23\linewidth]{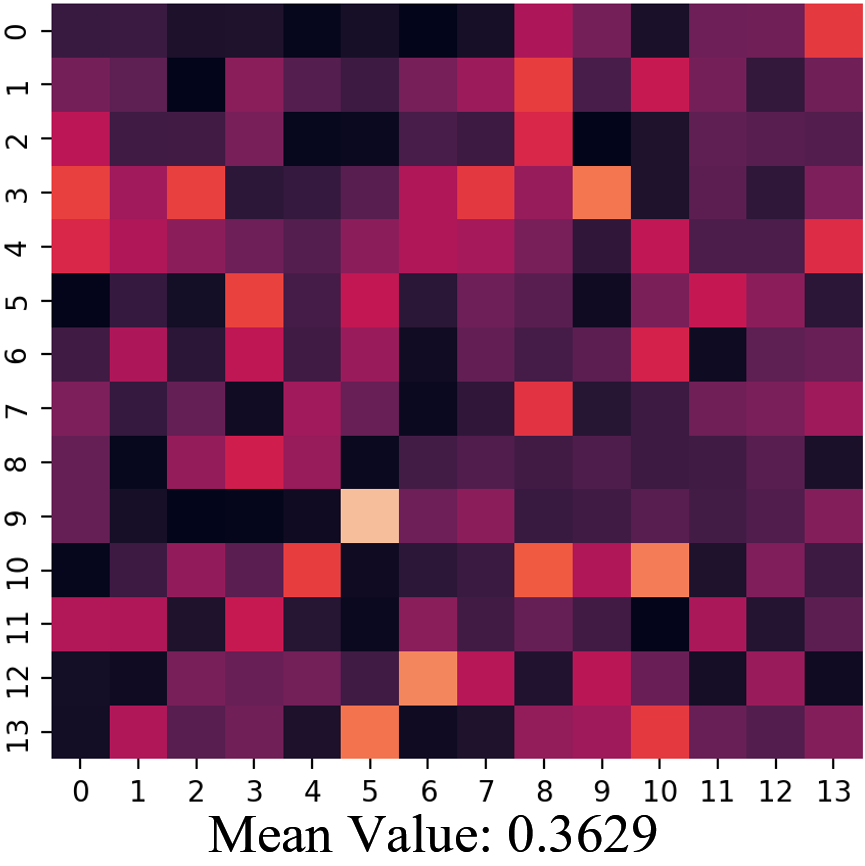}
}
\subfloat [\footnotesize Layer 2 ]{
\label{LAM2}
\includegraphics[width=0.23\linewidth]{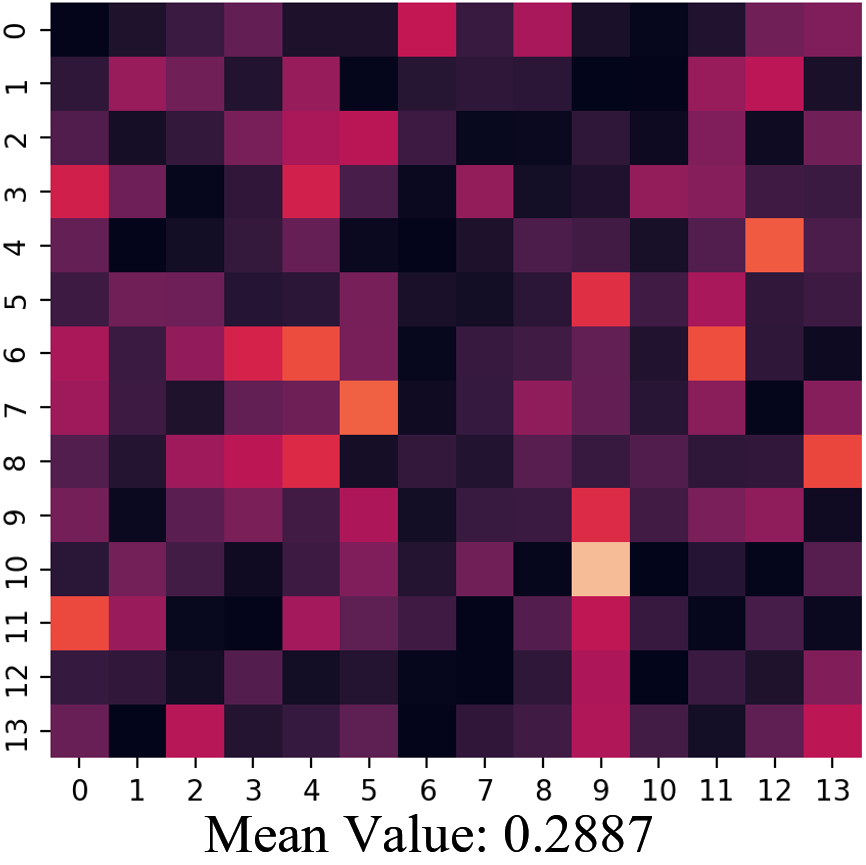}
}
\subfloat [\footnotesize Layer 3]{
\label{LAM3}
\includegraphics[width=0.23\linewidth]{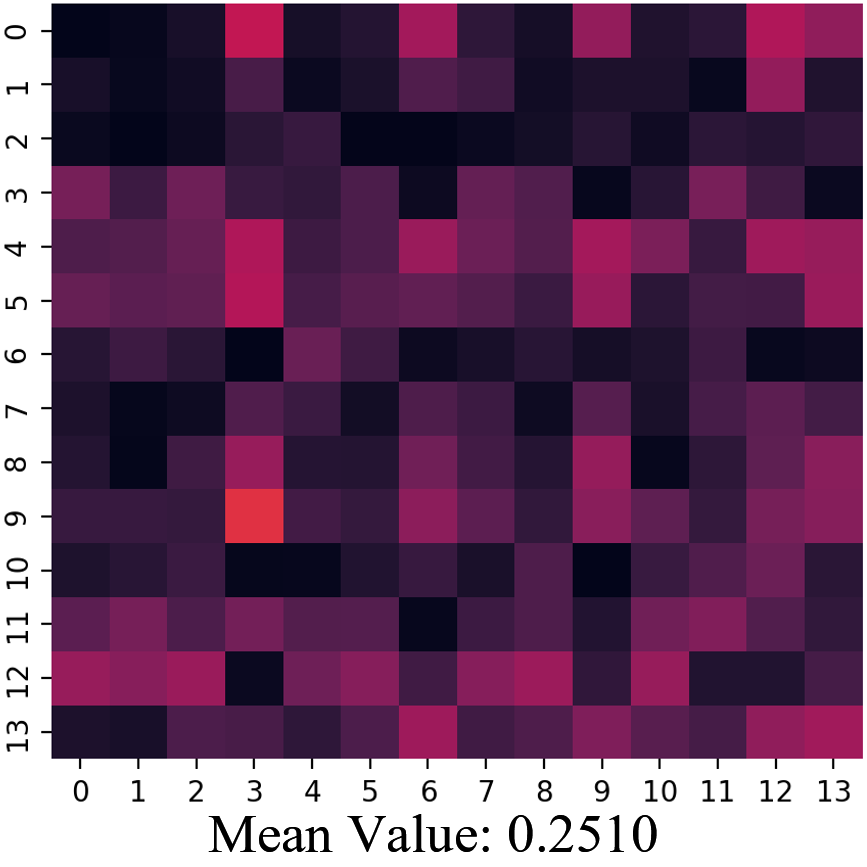}
}
\subfloat[\footnotesize Layer 4 ]{
\label{LAM4}
\includegraphics[width=0.272\linewidth]{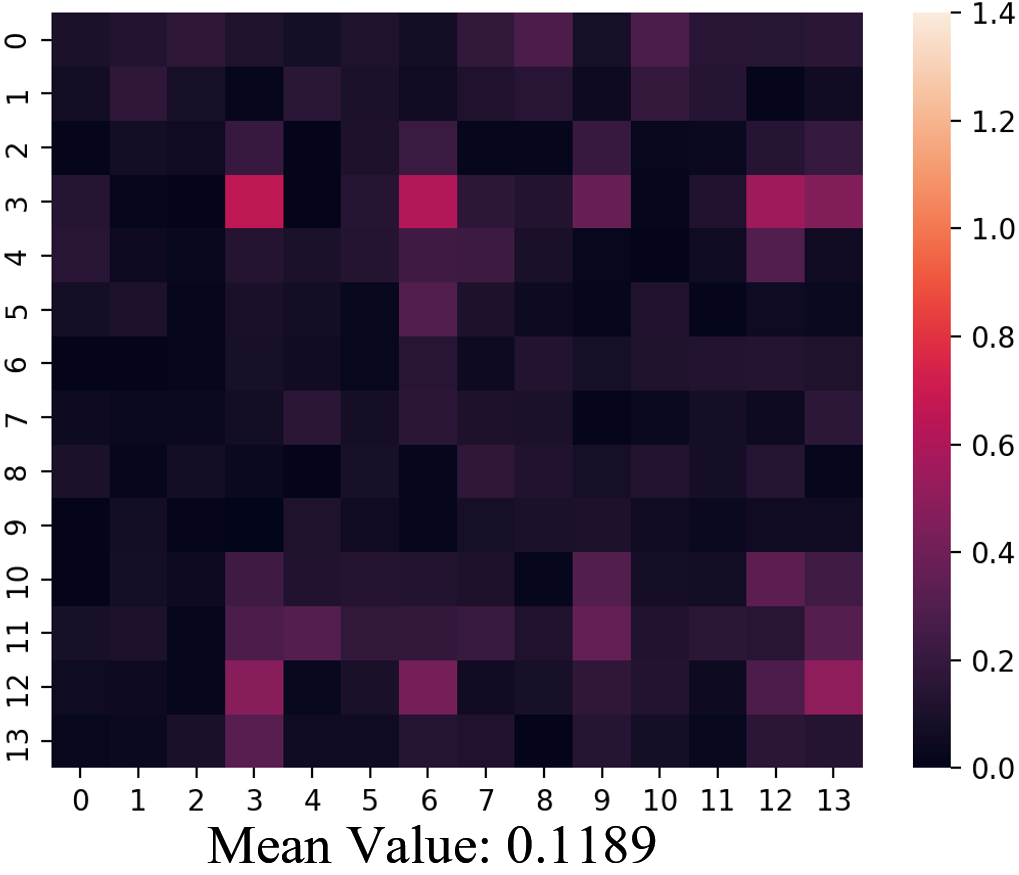}
}
\end{center}
\caption{The visualization results of LAM. (a) to (d) mean the LAMs in layers 1 to 4, respectively. }
\label{VLAM1}
\vspace{-5mm}
\end{figure*}

\begin{figure}
\setlength{\abovecaptionskip}{0pt}
\setlength{\belowcaptionskip}{0pt}
\begin{center}
\includegraphics[width=0.45\linewidth]{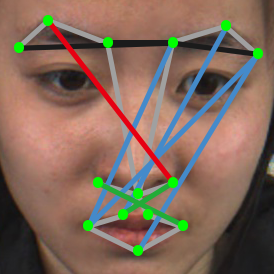}
\end{center}
   \caption{The visualization for LAM in layer 1. The line connecting two nodes indicates that the two nodes are strongly correlated, namely, the value in LAM is more lager. Red line: eyebrows-nose; Blue line: eyebrows-mouth; Green line: nose-mouth; Black line: eyebrows, where A-B indicates that the connects are between A and B regions.  }
\label{LAME}
\vspace{-5mm}
\end{figure}

To evaluate the effectiveness of LAM, the GTS-GN with LAM are compared with GTS-GN without LAM. Table \ref{AS} shows the comparison results. It turns out that both GTS-GN without and with AAU loss are improved after adding LAM. It demonstrates that LAM is effective for MER to learn a more reasonable adjacency matrix. Namely, LAM is superior to a fixed adjacency matrix.

\noindent\textbf{Visualization}: In order to illustrate the advantages of LAM intuitively, we visualize the learned LAM on CASME \uppercase\expandafter{\romannumeral2}. Fig. \ref{VLAM1} shows the heatmap of the learned LAMs in different layers. From this figure, the different layers have different LAMs, and as the layer increases, the average value of the learned LAM is smaller. It can be interpreted as that the node features in earlier layers represent low-level feature information, and the relationship between different nodes is strong to be aggregated to high-level feature information. Thus, in earlier layers, the relationship between different node features is modeled using an adjacency matrix with larger values. For the deeper layers, every node feature represents higher-level information and already interacts with other node features. Thus, an adjacency matrix with smaller values is enough to model the relationship between node features. Furthermore, we visualize the ten edges with larger values in the LAM of the last layer. As shown in Fig. \ref{LAME}, LAM can learn the relationship between different facial muscle regions. These regions do not have connections in the pre-defined adjacency matrix, \textit{e.g.} the relationship between the nodes feature of mouth and eyebrows regions. Thus, after introducing LAM, the model can aggregate the node features more effectively. 

Overall, LAM can consider the difference between different layers to build the relationship between different facial organs in multi-scales. In this way, more reasonable adjacency matrices can be learned to aggregate the node features.

\subsubsection{The Evaluation on Adaptive AU Loss}

\begin{table}
\setlength{\abovecaptionskip}{0pt}
\setlength{\belowcaptionskip}{0pt}
\vspace{0pt}
\caption{The ablation study on LAM and AAU Loss. ACC(\%).}
\begin{center}
\footnotesize
\begin{tabular}{|l|l|l|l|l|l|l|}
\hline
\multicolumn{3}{|c|}{Components}&\multicolumn{2}{c|}{CAMSE \uppercase\expandafter{\romannumeral2}}&\multicolumn{2}{c|}{SAMM} \\
\hline
GTS-GN&LAM&AAU Loss&ACC  &F1 &ACC  &F1\\
\hline\hline
 $\checkmark$ &$\times$ &$\times$&73.31& 0.717 &75.89  &0.741\\
 $\checkmark$ &$\checkmark$ &$\times$&74.90&0.732 &78.01 &\textbf{0.782}\\
$\checkmark$ &$\times$ &$\checkmark$&76.10&0.745 &78.01  &0.756\\
$\checkmark$&$\checkmark$&$\checkmark$&\textbf{77.29}& \textbf{0.765} & \textbf{79.43}  &\textbf{0.782}\\
\hline
\end{tabular}
\end{center}
\label{AS}
\vspace{-3mm}
\end{table}

\begin{table}
\setlength{\abovecaptionskip}{0pt}
\setlength{\belowcaptionskip}{0pt}
\vspace{0pt}
\caption{The evaluation on GTS-GN with AU loss. AU loss n denotes that AU loss constrain the features in the n-th layer. ACC(\%). }
\begin{center}
\footnotesize
\begin{tabular}{|l|p{20pt}|c|c|p{20pt}|c|}
\hline
\multirow{2}{*}{Methods} &\multicolumn{2}{c|}{CAMSE \uppercase\expandafter{\romannumeral2}} &\multicolumn{2}{c|}{SAMM} \\
\cline{2-5}
&ACC  &F1 &ACC  &F1\\
\hline\hline
GTS-GN                & 72.91& 0.697& 70.92& 0.673\\
GTS-GN with AU loss 1 & 73.31& 0.712& 75.18& 0.743\\
GTS-GN with AU loss 2 & 71.71& 0.697& 74.47& 0.713\\
GTS-GN with AU loss 3 & 73.31& 0.727& 75.18& 0.729\\
GTS-GN with AU loss 4 & 72.91& 0.708& 75.18& 0.724\\

\hline
\end{tabular}
\end{center}
\label{AULoss}
\vspace{-5mm}
\end{table}

In this section, we report the results of GTS-GN, GTS-GN with AU loss and AAU loss. In addition, AAU loss is evaluated in detail by visualization. 

First of all, we evaluate the AU loss in details. GTS-GN (layer 1) is taken as an example because it includes more constrained layers ($N_L = 4$). Table \ref{AULoss} reports the results that add AU loss to 4 layers, respectively. It turns out that constraining the features in different layers has different performance. Overall, the performance is improved after using AU loss. In addition, the performance maybe worse after adding AU loss in some layers, \textit{e.g.} layer 2 on CASME \uppercase\expandafter{\romannumeral2}. 

Next, AAU loss is effective and can improve the performance as shown in Table \ref{AS}. 
From Table \ref{AS}, both GTS-GN and GTS-GN with LAM achieve a better performance after adding AAU loss. It demonstrates that AAU loss is helpful to learn more discriminative features for recognizing MEs.

\begin{table}
\setlength{\abovecaptionskip}{0pt}
\setlength{\belowcaptionskip}{0pt}
\caption{The evaluation on AU loss and AAU Loss. ACC(\%).}
\centering
\footnotesize
\setlength{\tabcolsep}{3pt}
\subfloat[\footnotesize On CAMSE \uppercase\expandafter{\romannumeral2}]{
\begin{tabular}{|l|l|l|l|l|l|l|}
\hline
\multirow{2}{*}{Models} &\multicolumn{2}{c|}{Basic Model} &\multicolumn{2}{c|}{AU Loss}&\multicolumn{2}{c|}{AAU Loss} \\
\cline{2-7}
&ACC  &F1 &ACC &F1&ACC &F1\\
\hline\hline
GTS-GN (layer 1) & 72.91& 0.697&73.31& 0.728& \textbf{75.70}& \textbf{0.749}\\
GTS-GN (layer 2) & 74.10& 0.730&74.50& 0.738& \textbf{77.29}& \textbf{0.765}\\
GTS-GN (layer 3) & 74.90& 0.720&\textbf{75.30}& \textbf{0.745}& \textbf{75.30}& 0.744\\
GTS-GN (layer 4) & 70.52& 0.688&71.31& 0.697& -& -\\

\hline
\end{tabular}
       \label{AA:1}
}

\subfloat[\footnotesize On SAMM]{
\begin{tabular}{|l|l|l|l|l|l|l|}
\hline
\multirow{2}{*}{Models} &\multicolumn{2}{c|}{Basic Model} &\multicolumn{2}{c|}{AU Loss}&\multicolumn{2}{c|}{AAU Loss} \\
\cline{2-7}
&ACC  &F1 &ACC  &F1&ACC  &F1\\
\hline\hline
GTS-GN (layer 1) & 70.92& 0.673&75.18& 0.743& \textbf{77.30}& \textbf{0.765}\\
GTS-GN (layer 2) & 76.60& 0.739&75.18& 0.722& \textbf{79.43}& \textbf{0.778}\\
GTS-GN (layer 3) & 73.76& 0.700&77.30& 0.767& \textbf{78.72}& \textbf{0.782}\\
GTS-GN (layer 4) & 78.01& 0.782&77.30& 0.753& -& -\\
\hline
\end{tabular}
\label{GTS-GCN}
}
\label{AAULoss}
\vspace{-5mm}
\end{table}

Furthermore, as shown in Table \ref{AAULoss}, AAU loss is compared with AU loss and basic model (GTS-GN) under four types of GTS-GN. Since the constrained layer in GTS-GN (layer 4) only has one while $N_T>1$, GTS-GN (layer 4) with AAU loss don't have results. 
Compared with GTS-GN, AAU loss can greatly improve the performance in all cases. On the other hand, although AU loss doesn't promote GTS-GN (layer 2) on SAMM, it still improves the performances of the model in most cases. The above results show that before the classification layer, the use of AU information to constraint features can improve the performance of MER. More detailed, although AU loss and AAU loss have similar performance for GTS-GN (layer 3) on CASME \uppercase\expandafter{\romannumeral2}, for all other cases, AAU loss is superior to AU loss. It demonstrates that it is more advantageous to adaptively constrain multi-scale features in multiple layers than fixedly constrain the features of a certain layer.
\begin{figure*}
\begin{center}
\vspace{-2mm}
\setlength{\abovecaptionskip}{0pt}
\setlength{\belowcaptionskip}{0pt}
\subfloat[Layer 1]{
\label{NoLoss1}
\includegraphics[width=0.25\linewidth]{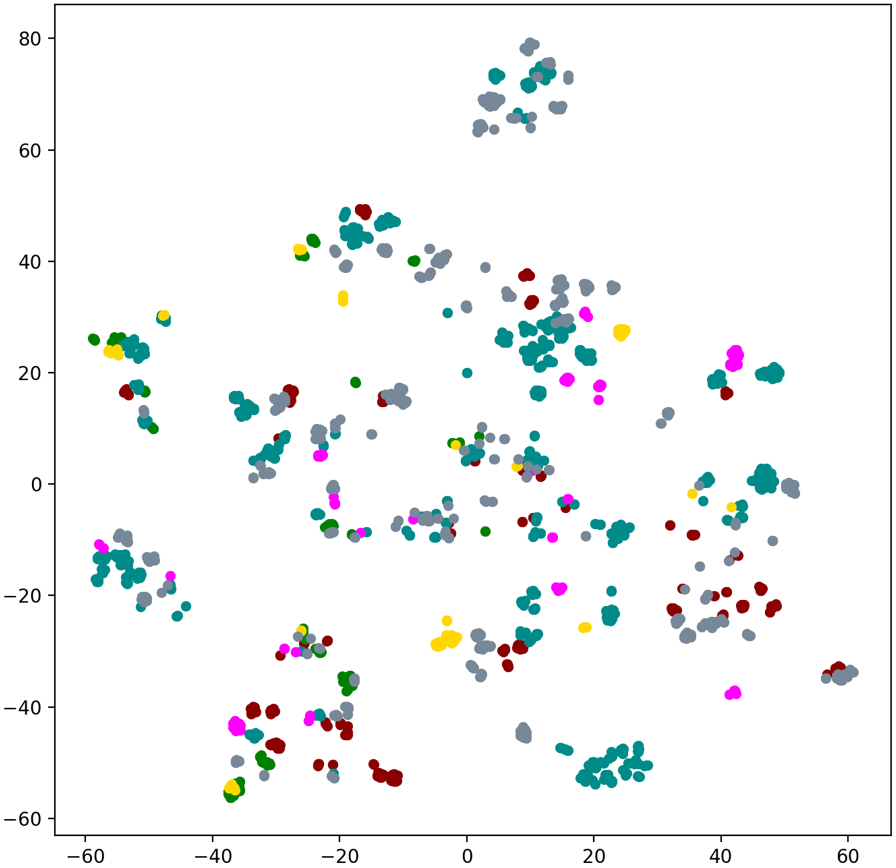}
}
\subfloat [ Layer 2]{
\label{NoLoss2}
\includegraphics[width=0.25\linewidth]{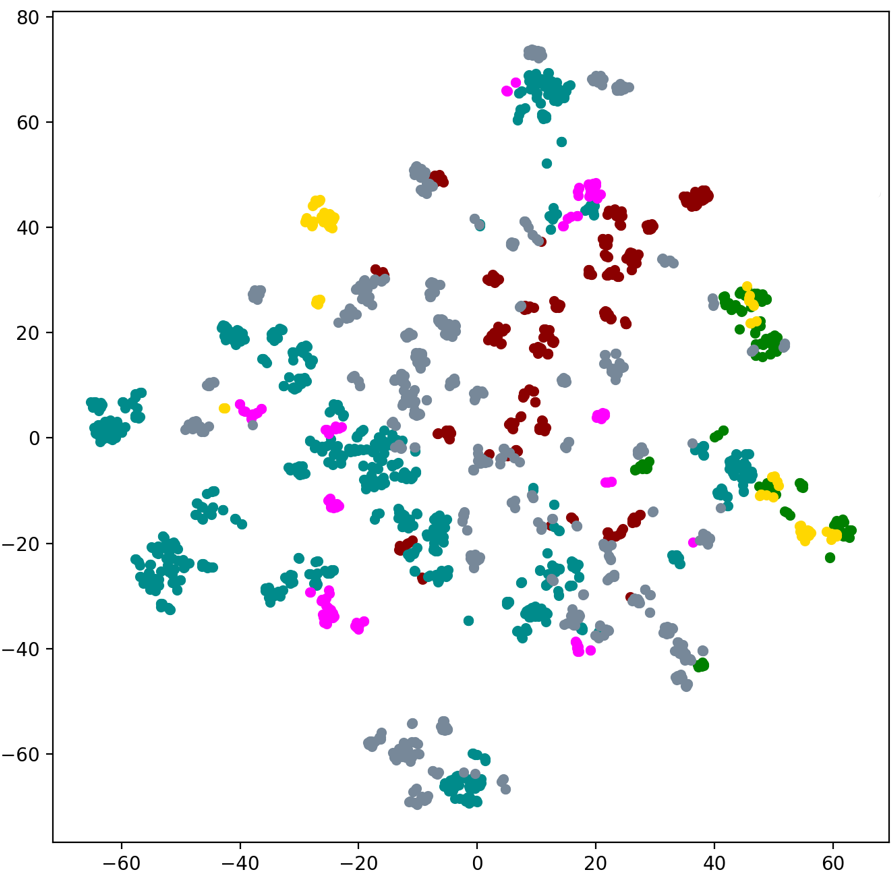}
}
\subfloat [Layer 3]{
\label{NoLoss3}
\includegraphics[width=0.25\linewidth]{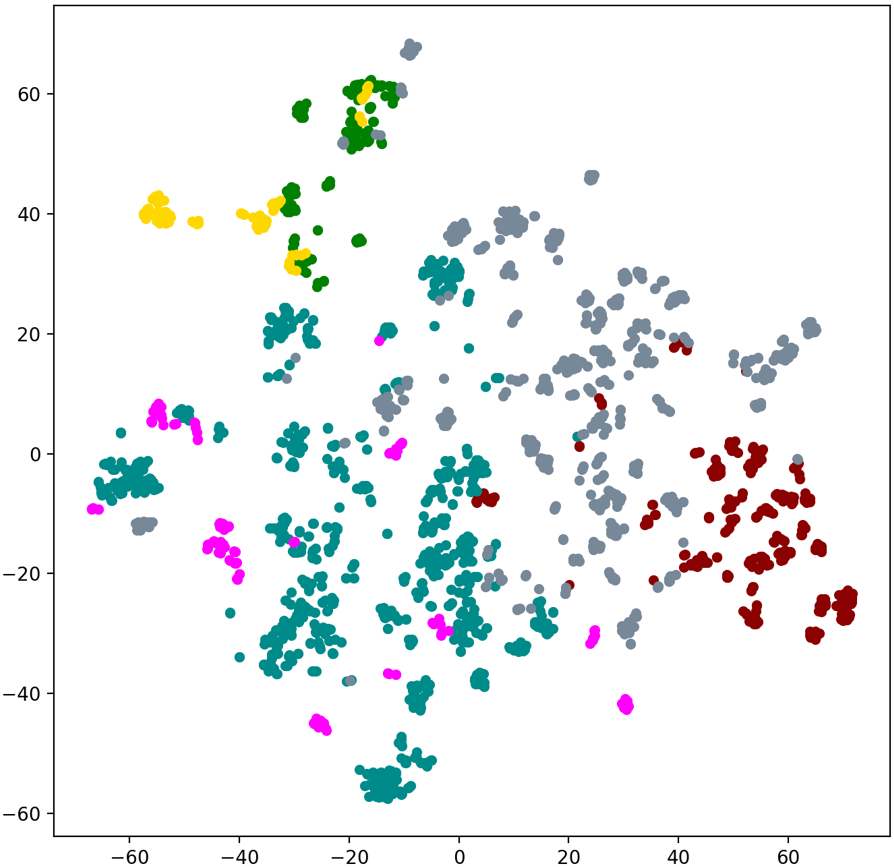}
}
\subfloat [ Layer 4]{
\label{NoLoss4}
\includegraphics[width=0.25\linewidth]{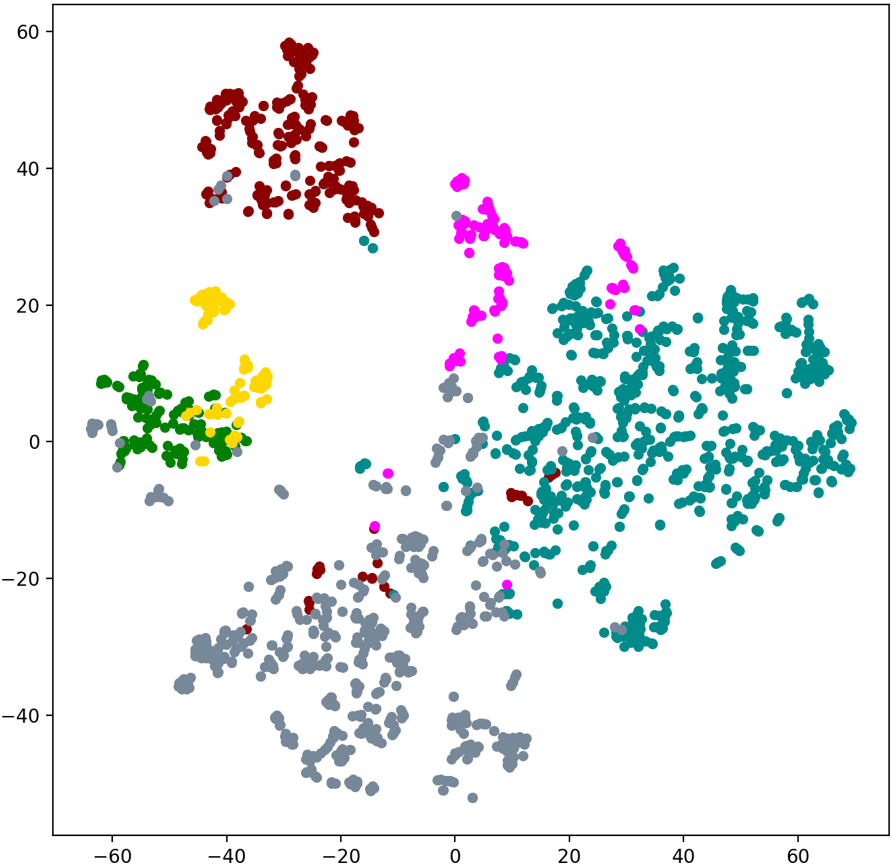}
}

\vspace{-2mm}
\subfloat [Layer 1]{
\label{Loss1}
\includegraphics[width=0.25\linewidth]{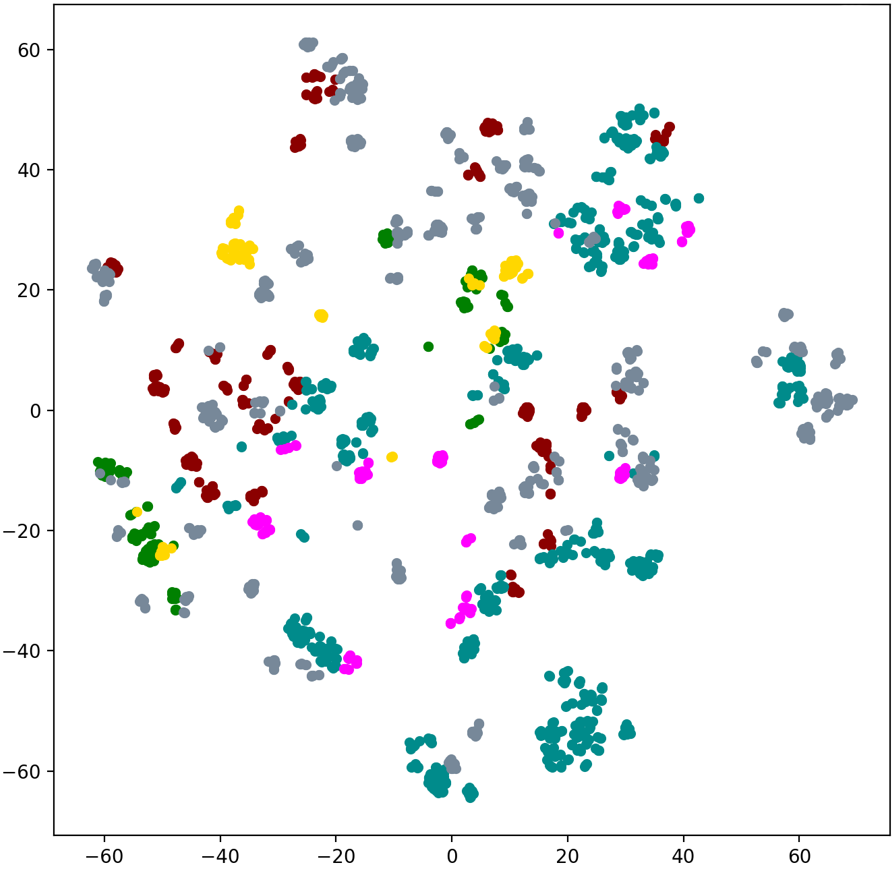}
}
\subfloat [Layer 2 ]{
\label{Loss2}
\includegraphics[width=0.25\linewidth]{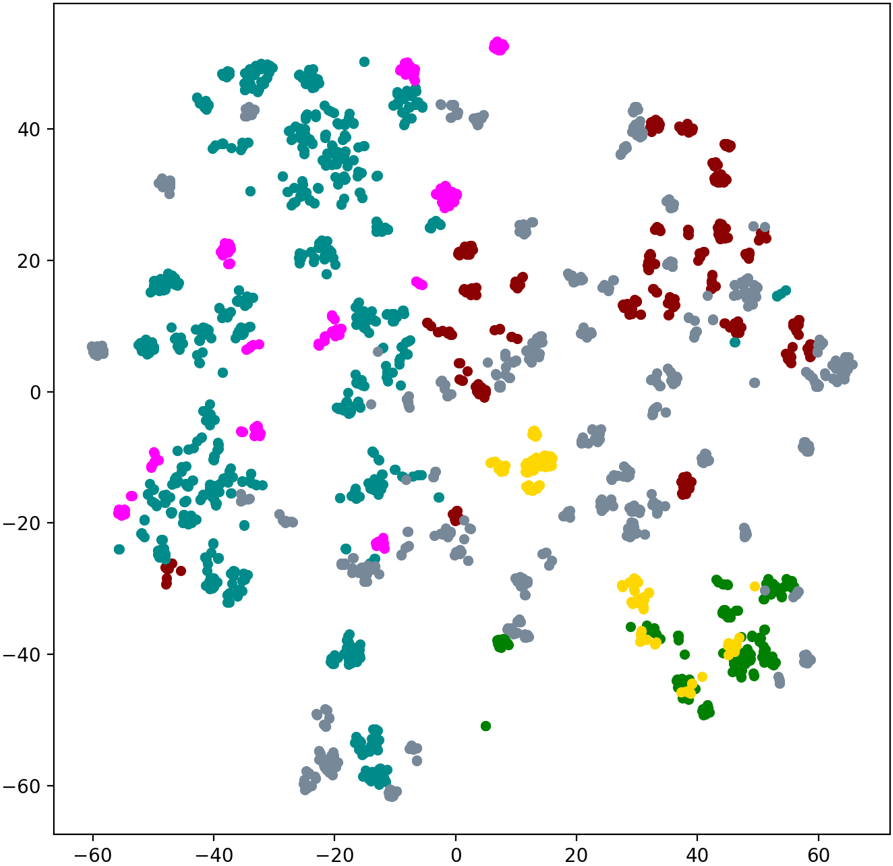}
}
\subfloat [Layer 3]{
\label{Loss3}
\includegraphics[width=0.25\linewidth]{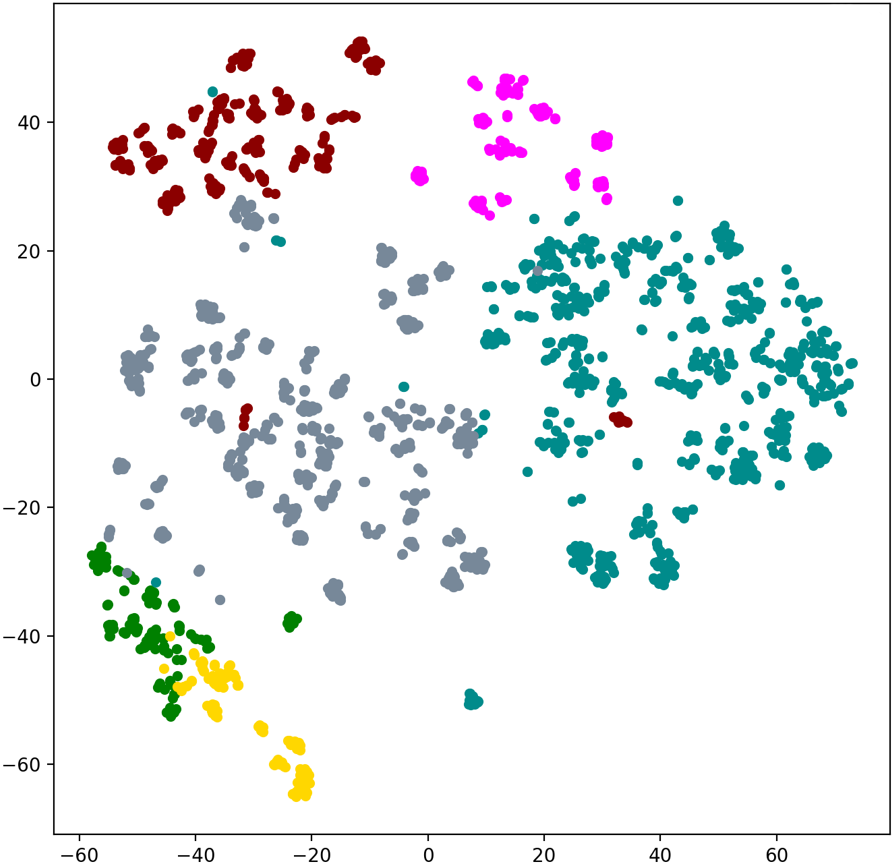}
}
\subfloat [ Layer 4]{
\label{Loss4}
\includegraphics[width=0.25\linewidth]{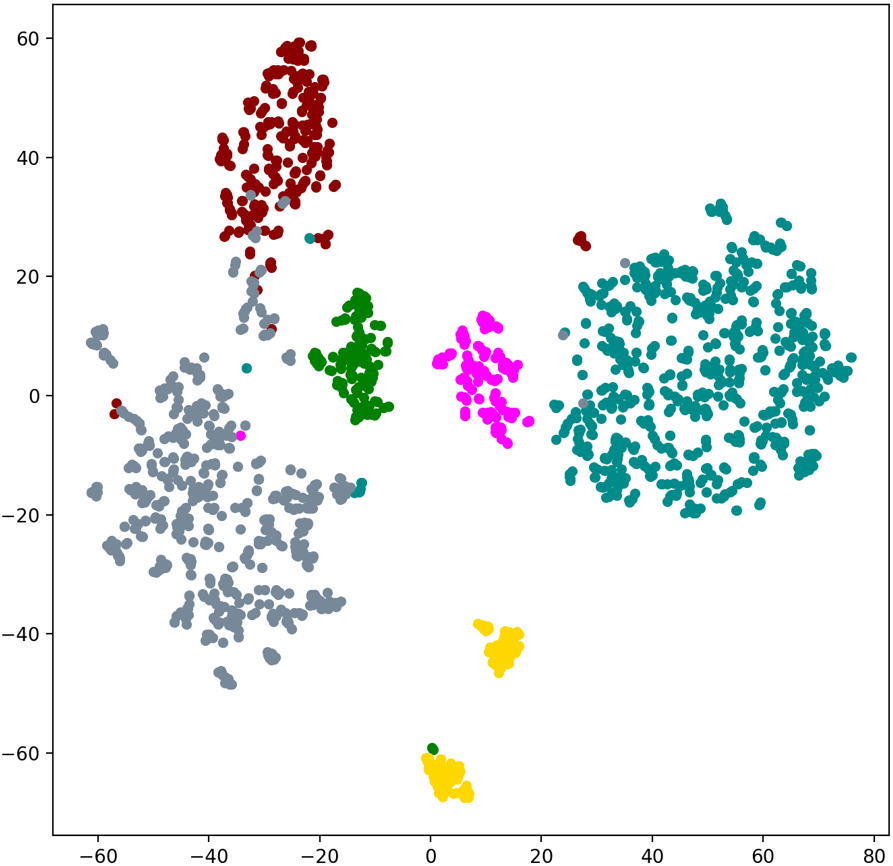}
}
\vspace{-1mm}

\subfloat{
\label{indicator}
\includegraphics[width=0.2\linewidth]{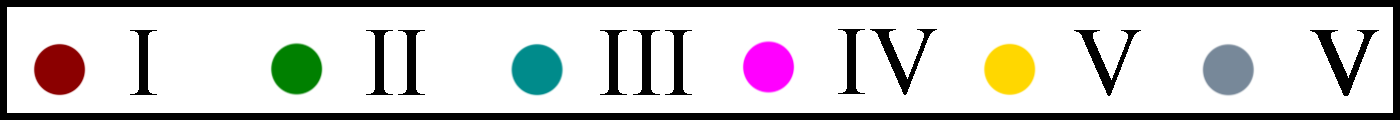}
}

\vspace{-3mm}
\end{center}
\caption{The visualization of the features in different layers using t-SNE. The first and second rows are the results without and with AAU loss, respectively. }
\label{LossV}
\vspace{-3mm}
\end{figure*}

\noindent\textbf{Visualization}: We further analyze AAU loss by showing the learned weight values $\frac{W_r^2}{\sum_{r=1}^{N_L} W_r^2}$ and visualizing the learned features in different layers. GTS-GN (layer 1) is taken as an example, so there are four learnable weights, namely $N_L=4$. Table \ref{LW} shows the learned four weights for 26 subjects on CASME \uppercase\expandafter{\romannumeral2}. It turns out that the two weights for the first two layers are much larger than that for the last two layers. Namely, AAU loss focuses on limiting the features in the first two layers to aggregate AU information. In fact, the movements of facial landmarks have a direct relationship with AUs. Thus, AAU loss can constrain the features to achieve the aggregation of high-level AU information in more early layers. 

To verify above points, we further visualize the features in different layers. t-SNE\cite{T-SNE} is employed to visualize these features to be a scatter plot. The visualization results with and without AAU loss are shown in Fig. \ref{LossV}. Obviously, comparing with not using AAU loss, the extracted features have some regularities on the first two layers after using AAU loss. Specifically, for the first layer, the features corresponding to the same category hardly appear cluster when not using AAU loss. While after using AAU loss, those begin to cluster. For the second layer, the clustering effect using AAU loss is better than that not using AAU loss. For instance, for category III, the features not using AAU loss are distributed on the whole plane, while those using AAU loss are mainly distributed on the left side. In addition, Fig. \ref{Loss2} shows that the features using AAU loss have some special intersection relationships that correspond to the relationships between AUs and MEs. In fact, a ME is related to single or multiple AUs, and different MEs may have a common AU, \textit{e.g.} both II and V include AU1, and both III and IV include AU4,AU5 and AU7 (Corresponding to Fig. \ref{Loss2}, the features of II and V or III and IV have some intersections). Above results demonstrate that AAU loss can constrain the features in the earlier layers to represent high-level AU information, which is consistent with the learned weights. Based on the AU-related features learned in the first two layers, AAU has obvious advantages for the learned features in the latter two layers. Specifically, for the third layer as shown in Fig. \ref{NoLoss3} and \ref{Loss3}, the features using AAU loss obviously are superior to those not using AAU loss whether in terms of intra-class or inter-class. For the last layer, the features using AAU loss still maintain the advantage with larger inter-class and smaller intra-class. 

Overall, in earlier layers, the AAU loss-constrained model focuses on learning high-level AU features from facial landmarks, while in deeper layers, it focuses on learning high-level ME features from high-level AU features. By introducing AAU loss, the learned features in the earlier layers have some rules derived from AU information. Based on it, the learned features in the deeper layers are more discriminative to classify MEs.

\begin{table}
\setlength{\abovecaptionskip}{0pt}
\setlength{\belowcaptionskip}{0pt}
\caption{The learned weights for 26 subjects on CASME \uppercase\expandafter{\romannumeral2}.}
\begin{center}
\footnotesize
\begin{tabular}{|l|p{20pt}|c|c|p{20pt}|c|}
\hline
layer & 1 &2 & 3 &4\\
\hline
Subject 1& 0.5662 & 0.4127 & 0.0010 & 0.0201\\
Subject 2& 0.6992 & 0.2246 & 0.0118 & 0.0643\\
Subject 3& 0.6732 & 0.2727 & 0.0064 & 0.0477\\
Subject 4& 0.5312 & 0.4448 & 0.0055 & 0.0119\\
Subject 5& 0.6153 & 0.3296 & 0.0005 & 0.0546\\
Subject 6& 0.6404 & 0.3064 & 0.0003 & 0.0529\\
Subject 7& 0.6879 & 0.2514 & 0.0111 & 0.0497\\
Subject 8& 0.6078 & 0.3593 & 0.0000 & 0.0329\\
Subject 9& 0.6376 & 0.2848 & 0.0002 & 0.0774\\
Subject 10& 0.6724 & 0.2545 & 0.0025 & 0.0705\\
Subject 11& 0.6411 & 0.3087 & 0.0005 & 0.0498\\
Subject 12& 0.6625 & 0.2844 & 0.0033 & 0.0499\\
Subject 13& 0.6161 & 0.3341 & 0.0002 & 0.0495\\
Subject 14& 0.6056 & 0.3657 & 0.0000 & 0.0287\\
Subject 15& 0.6035 & 0.3484 & 0.0009 & 0.0417\\
Subject 16& 0.6801 & 0.2621 & 0.0080 & 0.0498\\
Subject 17& 0.5765 & 0.3853 & 0.0025 & 0.0357\\
Subject 18& 0.6387 & 0.3238 & 0.0017 & 0.0358\\
Subject 19& 0.6515 & 0.3021 & 0.0022 & 0.0441\\
Subject 20& 0.6758 & 0.2470 & 0.0027 & 0.0745\\
Subject 21& 0.5488 & 0.4267 & 0.0033 & 0.0211\\
Subject 22& 0.6039 & 0.3501 & 0.0007 & 0.0452\\
Subject 23& 0.6418 & 0.3126 & 0.0009 & 0.0446\\
Subject 24& 0.6341 & 0.3256 & 0.0007 & 0.0396\\
Subject 25& 0.6668 & 0.2760 & 0.0035 & 0.0537\\
Subject 26& 0.6781 & 0.2558 & 0.0050 & 0.0611\\
Average   & 0.6329 & 0.3172 & 0.0029 & 0.0464\\
\hline
\end{tabular}
\end{center}
\label{LW}
\vspace{-3mm}
\end{table}

\begin{table}
\setlength{\abovecaptionskip}{0pt}
\setlength{\belowcaptionskip}{0pt}
\vspace{0pt}
\caption{The evaluation on three types of landmark point sets. ACC(\%). }
\begin{center}
\footnotesize
\begin{tabular}{|l|p{20pt}|c|c|p{20pt}|c|}
\hline
\multirow{2}{*}{Point number} &\multicolumn{2}{c|}{CAMSE \uppercase\expandafter{\romannumeral2}} &\multicolumn{2}{c|}{SAMM} \\
\cline{2-5}
&ACC  &F1 &ACC  &F1\\
\hline\hline
14 & 72.91& 0.716& 75.17& 0.732\\
31 & 72.91& 0.693& 72.34& 0.691\\
68 & 72.50& 0.704& 68.79& 0.608\\
\hline
\end{tabular}
\end{center}
\label{RPoint}
\vspace{-5mm}
\end{table}

\subsection{Parameter Evaluation}

In this section, the affect of the used landmark points and the trade-off parameter $\beta$ are evaluated for the proposed method. We test three types of the landmark point sets as shown in Fig. \ref{point}, and $\beta$ are tested to show the balance between AAU loss and ME loss.

\noindent\textbf{The Affect of the Used Landmark Points}:

\begin{figure}
\setlength{\abovecaptionskip}{0pt}
\setlength{\belowcaptionskip}{0pt}
\begin{center}
\includegraphics[width=0.8\linewidth]{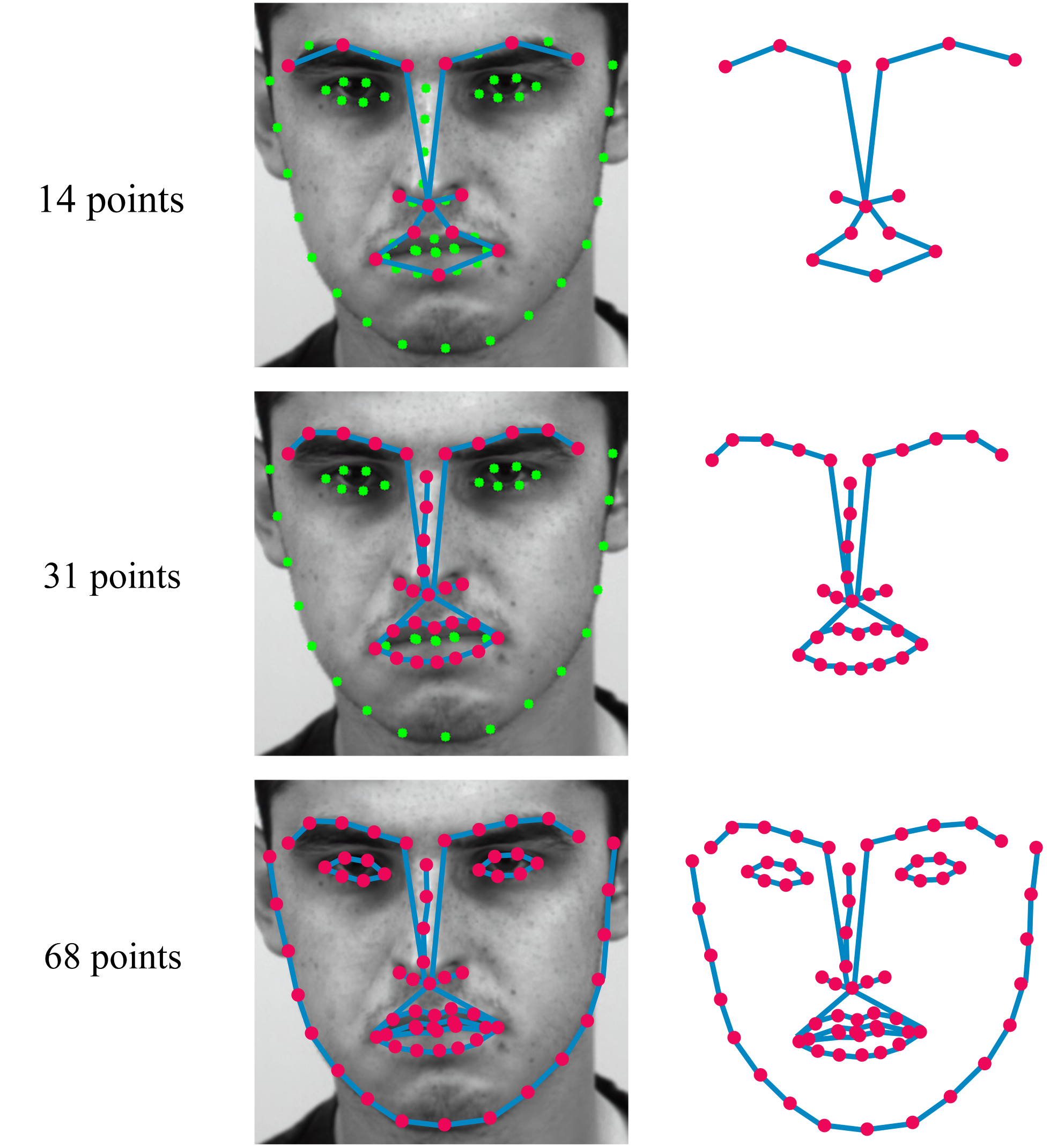}
\end{center}
   \caption{Three types of landmark point sets. }
\label{point}
\vspace{-5mm}
\end{figure}

Except for the 14 points of the graph defined in the 3.1 section, other two types of landmark point sets are tested, including 1) 31 points: all landmark points of mouth, nose and eyebrows regions; and 2) 68 points: all landmark points of whole face. Table \ref{RPoint} reports the comparative results. Obviously, 14 points set has an obvious advantage and the performance of 68 points set is worst. It turns out that under including the key information of eyebrows, nodes and mouth, with the point number increases, the performance drops. Compared with 14 points set, 31 points set includes redundant information. 68 points set not only contains a lot of redundant information, but also some interference information, \textit{e.g.} the eyes and facial contours. Also, 14 points set has fewer points, which reduces the computational cost and improves efficiency. Thus, removing interference points as well as redundant points and retaining key information for eyebrows, nose, and mouth is beneficial to improve performance and efficiency.

\begin{figure}
\vspace{-5mm}
\setlength{\abovecaptionskip}{0pt}
\setlength{\belowcaptionskip}{0pt}
\begin{center}
\includegraphics[width=0.9\linewidth]{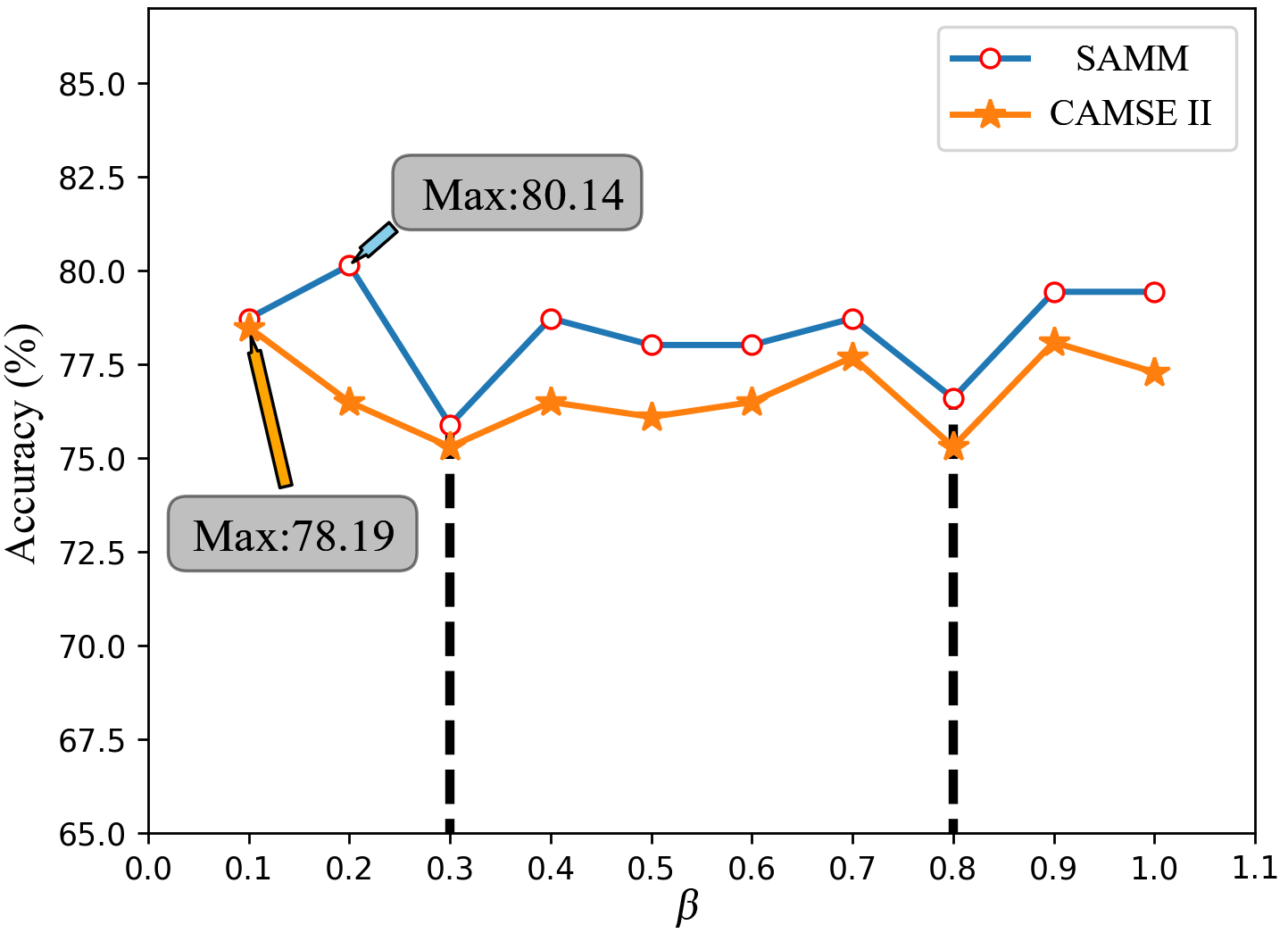}
\end{center}
   \caption{The evaluation on $\beta$. The black dotted line represents the valley value. }
\label{alpha}
\vspace{-5mm}
\end{figure}

\noindent\textbf{The Affect of $\beta$}: 
Fig. \ref{alpha} shows the evaluation results of $\beta$. The proposed method achieves the highest accuracy under $\beta$=0.1 (78.49\%) on CAMSE \uppercase\expandafter{\romannumeral2} and $\beta$=0.2 (80.14\%) on SAMM. Furthermore, under $\beta$=0.3 and 0.8, the performances on both datasets reach the valley. In general, the performance under $\beta$ closing to 0.1 or 1 is superior to that under $\beta$ in the middle value from 0.1 to 1. In detail, under $\beta$ closing to 0.1, the highest accuracy can be achieved. It means that it is a better choice to take AU information as auxiliary information and remain the maintain dominance of ME loss.

\subsection{Comparing with Other Methods}

To evaluate the proposed method, we compare to some existing state-of-the-art methods. As shown in Table \ref{O5}, comparing with the existing methods recognizing five ME categories based on AU labels, GTS-GN provides a much better performance on both datasets. GTS-GN achieves a new SOTA performance, especially, in terms of ACC and F1-score, which are 11.53\% and 0.198 higher on CASME \uppercase\expandafter{\romannumeral2}, and 1.47\% and 0.169 higher on SAMM than TL.

For a more comprehensive comparison, we further test the results with five ME categories based on self-report labels. This is also the mainstream test categories. According to Table \ref{E5}, GTS-GN gets a comparable performance as a whole comparing with recent methods. Specifically, compared to the CNN-based methods, GEME and LGCcon, GTS-GN provides a much better performance on both datasets. Also, GACN and GRAUF are SOTA methods. Comparing with GRAUF, GTS-GN get a bit worse results on SAMM and a better results on CASME \uppercase\expandafter{\romannumeral2}, especially in terms of F1-score (0.107 higher than GRAUF on CASME \uppercase\expandafter{\romannumeral2}). Different from our method with a compact landmarks representation, GRAUF aggregated more information, \textit{e.g.} AUs and the magnified shape. Also, GRAUF needs to employed several models to deal with different features, which increases model complexity and reduces efficiency. Comparing with GACN, the proposed method also consistenty shows obvious advantages on CASME \uppercase\expandafter{\romannumeral2}, but on SAMM, GACN get better performances. This maybe because that except geometric features of landmarks, GACN also utilized optical flow map for dynamic information. Thus, GACN needs to spend much more computational cost to extract optical flow.

Overall, different from these existing methods, the compact landmarks input makes the proposed method achieving a competitive performance with a lot less computational cost. All results demonstrate the proposed method is effective and more suitable for practical applications.

\begin{table}
\setlength{\abovecaptionskip}{0pt}
\setlength{\belowcaptionskip}{0pt}
\vspace{0pt}
\caption{Comparing with other methods under 5 categories based on AUs annotation. ACC(\%).}
\begin{center}
\footnotesize
\begin{tabular}{|l|p{20pt}|c|c|p{20pt}|c|}
\hline
\multirow{2}{*}{Methods} &\multicolumn{2}{c|}{CAMSE \uppercase\expandafter{\romannumeral2}} &\multicolumn{2}{c|}{SAMM} \\
\cline{2-5}
&ACC  &F1 &ACC  &F1\\
\hline\hline
LBP-TOP\cite{ObClass-Davison2017}&67.80& 0.510 &44.70  & 0.350\\
HOOF\cite{ObClass-Davison2017}&69.64& 0.560 & 42.17  & 0.330\\
HOG3D\cite{ObClass-Davison2017} &69.53 &0.510&34.16 & 0.220\\
TL\cite{ObClass-TL-Peng} & 75.68 & 0.650& 70.59 & 0.540\\
ELRCN-TE\cite{ELTRCN-Khor} &52.44 &0.500&N/A&N/A\\
The proposed method & \textbf{87.21}&\textbf{0.848}&\textbf{72.06}&\textbf{0.709}\\
\hline
\end{tabular}
\end{center}
\label{O5}
\vspace{-5mm}
\end{table}

\begin{table}
\large
\setlength{\abovecaptionskip}{0pt}
\setlength{\belowcaptionskip}{0pt}
\vspace{0pt}
\caption{Comparing with other methods under 5 categories based on self-report annotation. ACC(\%).}
\begin{center}
\footnotesize
\begin{tabular}{|l|p{20pt}|c|c|p{20pt}|c|}
\hline
\multirow{2}{*}{Methods} &\multicolumn{2}{c|}{CAMSE \uppercase\expandafter{\romannumeral2}} &\multicolumn{2}{c|}{SAMM} \\
\cline{2-5}
&ACC  &F1 &ACC  &F1\\
\hline\hline
Hierarchical LBP-IP~\cite{KGSL-Zong} &63.83&0.61  &N/A &N/A\\
Discriminative LBP-IP~\cite{DLBP-RIP-Huang}&64.78 &N/A  &N/A &N/A\\
Sparse MDMO\cite{SparseMDMO-Liu}& 66.95& 0.691& N/A& N/A\\
LBP-SDG~\cite{Wei}&71.32& 0.665&N/A& N/A\\
KTGSL~\cite{Wei-KTGSL}&72.58& 0.682&56.11& 0.493\\
SSSN\cite{DSSN-Khor}& 71.19& 0.715& 56.62& 0.451\\
DSSN\cite{DSSN-Khor}& 70.78& 0.729& 57.35& 0.464\\
G-TCN\cite{G-TCN} & 73.98& 0.725& 75.00& 0.699\\
GEME\cite{GEME} & 75.20& 0.735& 55.88& 0.454\\
LGCcon\cite{LGCcon-Yante} & 65.02& 0.640& 40.90& 0.340\\
GRAUF\cite{Lei2} & 74.27& 0.705& 74.26& 0.705\\
GACN\cite{GACNN}& 81.30 & 0.709 &\textbf{ 88.24}& \textbf{ 0.828}\\
The proposed method &\textbf{81.78}& \textbf{0.812}& 71.32&0.711\\
\hline
\end{tabular}
\end{center}
\label{E5}
\vspace{-5mm}
\end{table}

\section{Conclusion}

This paper explored the contribution of facial landmarks and demonstrated the effectiveness of facial landmarks for MER. Notably, only the geometric information of facial landmarks is aggregated by the graph model to achieve MER task. 
We first customized a GM-Graph based on the facial landmarks of three key frames to model the geometric and dynamic information in ME videos. Then, SS module was proposed to learn the deep spatial and temporal features of GM-Graph. The experimental results demonstrate that SS module can aggregate the spatial and temporal information better. Namely, it's more suitable to introduce both GCN and TCN to separately aggregate the spatial and temporal information. To further improve the performance, geometric information including the low-order coordinates and high-order semantic features are both involved. Therefore, a new graph model GTS-GN was proposed, which models information interaction and takes better use of complementary information from two types of geometric features. Furthermore, LAM can automatically learn a more reasonable graph structure that builds the relationship between different facial muscle regions and between different nodes. In addition, AAU loss can reasonably learn and build strong correlations between facial landmarks, AUs and MEs. AAU loss can adaptively constrain the multi-scale movement features to aggregate AU information with an efficient way, learning more discriminative ME features. 

This work encourages further investigation of this framework that takes low-dimensional landmarks as input to extract compact geometric features with the graph-based model. This framework is more valuable for practical applications due to its low computational costs and at the same time comparable and even superior performance.


\section*{Acknowledgment}

This work was partly supported by the Postgraduate Research and Practice Innovation Program of Jiangsu Province (Grant KYCX19$\_$0899), partly by the National Natural Science Foundation of China (NSFC) under Grants 72074038, 61971236 and 62076122.

\bibliography{mybibfile}

\bibliographystyle{IEEEtran}
\vspace{-15mm}

%

\begin{IEEEbiography}[{\includegraphics[width=1in,height=1.25in]{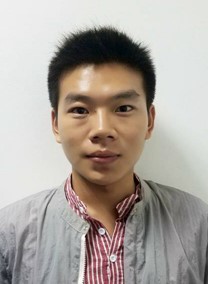}}]{Jinsheng Wei}
is currently a lecturer at the Nanjing University
of Posts and Telecommunications (NUPT), Nan-
jing, China. He received his Ph.D. degree from NUPT, China, in 2022. From 2020 to 2021, he was a Visiting Student with the Center for Machine Vision and Signal Analysis, University of Oulu, Finland. His current research interests include micro-expression recognition, image processing, pattern recognition, computer vision, and machine learning.
\vspace{-15mm}
\end{IEEEbiography}
\begin{IEEEbiography}[{\includegraphics[width=1in,height=1.25in]{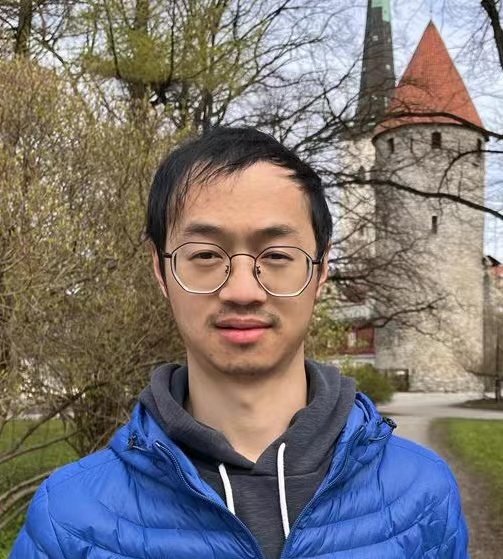}}]{Wei Peng}
is currently Postdoctoral Scholar at Stanford University. He received his Ph.D. degree from University of Oulu, Finland, in 2022. He received the M.S. degree in computer science from the Xiamen University, Xiamen, China, in 2016. His articles have published in mainstream conferences and journals, such as CVPR, AAAI, ICCV, ACM Multimedia, TPAMI, and TIP. His current research interests include machine learning, medical imaging analysis, and neuroscience.
\vspace{-25mm}
\end{IEEEbiography}
\begin{IEEEbiography}[{\includegraphics[width=1in,height=1.25in]{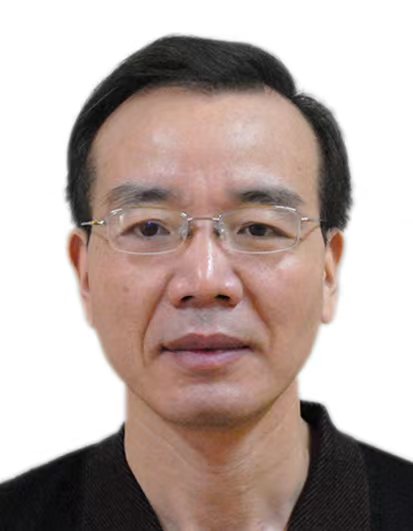}}]{Guanming Lu}
received the B.E. degree in radio engineering and the M.S. degree in communication and electronic systems from the Nanjing University of Posts and Telecommunications (NUPT), Nanjing, China, in 1985 and 1988, respectively, and the Ph.D. degree in communication and information systems from Shanghai Jiao Tong University, Shanghai, China, in 1999. He is currently a Professor with the College of Communication and Information Engineering, NUPT. His current research interests include image processing, affective computing, and machine learning. (Address: College of Telecommunications and Information Engineering, Nanjing University of Posts and Telecommunications, Nanjing 210003 China; Email: lugm@njupt.edu.cn)
\vspace{-25mm}
\end{IEEEbiography}
\begin{IEEEbiography}[{\includegraphics[width=1in,height=1.25in]{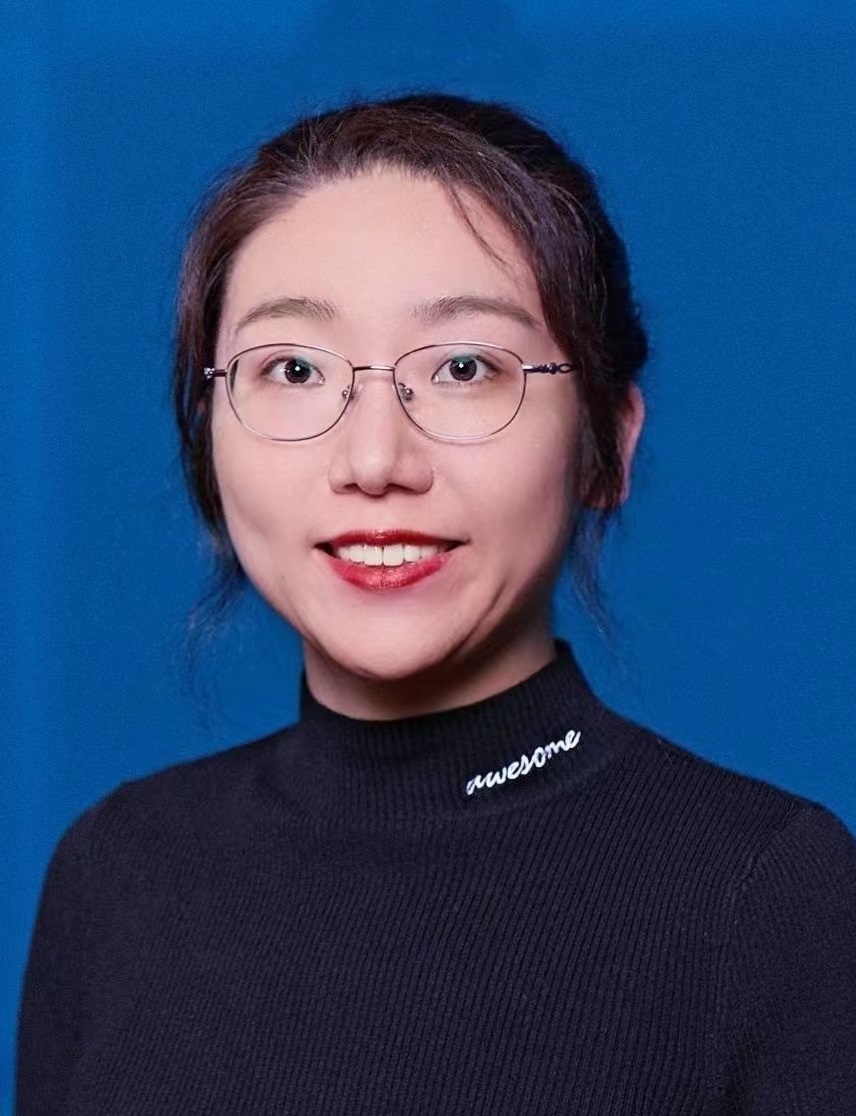}}]{Yante Li}
 received her Ph.D. degree in Computer
Science from University of Oulu, Oulu, Finland,
in 2022. She is currently a Postdoc in the Center
for Machine Vision and Signal Analysis of University of Oulu. Her current research interests
include affective computing, micro-expression
analysis and facial action unit detection.
\vspace{-25mm}
\end{IEEEbiography}
\begin{IEEEbiography}[{\includegraphics[width=1in,height=1.25in]{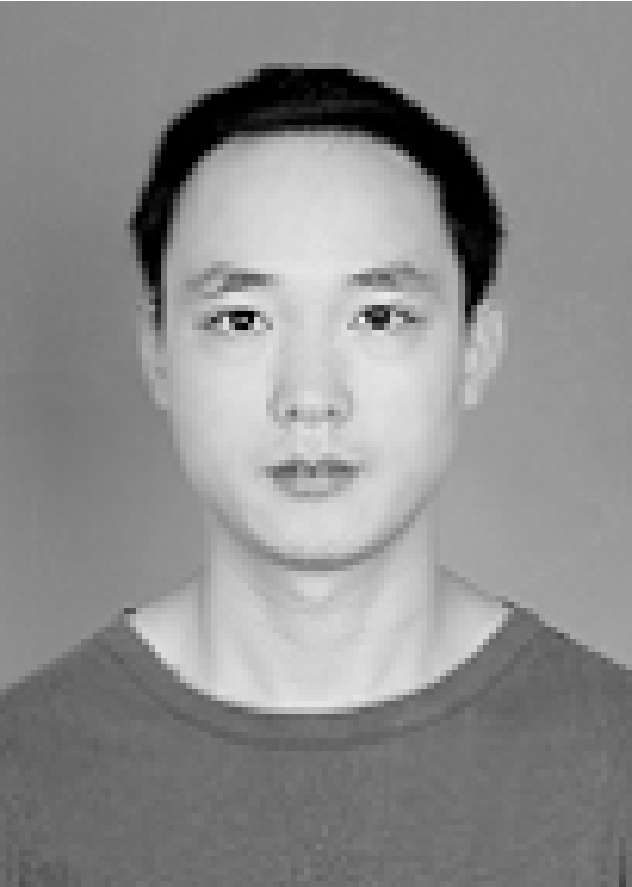}}]{Jingjie Yan}
received the B.E. degree in Electronic Science and Technology in 2006 and the M.S. degree in Signal and Information Processing in 2009, both from China University of Mining and Technology, and a Ph.D. degree in Single and Information Processing from Southeast University, Nanjing, China, in 2014. He is currently a Associate Professor in College of Communication and Information Engineering, Nanjing University of Posts and Telecommunications (NUPT). His current research interests include pattern recognition, affective computing, computer vision and machine learning.
\vspace{-25mm}
\end{IEEEbiography}
\begin{IEEEbiography}[{\includegraphics[width=1.1in,height=1.2in]{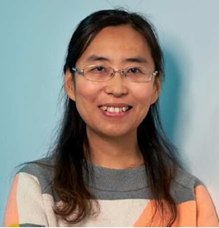}}]{Guoying Zhao (IEEE Fellow 2022) }
received the Ph.D. degree in computer science from the Chinese Academy of Sciences, Beijing, China, in 2005. She is currently an Academy Professor and full Professor (tenured in 2017) with University of Oulu. She is also a visiting professor with Aalto University. She is a member of Finnish Academy of Sciences and Letters, IEEE Fellow, IAPR Fellow and AAIA Fellow. She has authored or co-authored more than 300 papers in journals and conferences with 20800+ citations in Google Scholar and h-index 67. She is panel chair for FG 2023, publicity chair of 22nd Scandinavian Conference on Image Analysis (SCIA 2023), was co-program chair for ACM International Conference on Multimodal Interaction (ICMI 2021), co-publicity chair for FG2018, and has served as area chairs for several conferences and was/is associate editor for IEEE Trans. on Multimedia, Pattern Recognition, IEEE Trans. on Circuits and Systems for Video Technology, and Image and Vision Computing Journals. Her current research interests include image and video descriptors, facial-expression and micro-expression recognition, emotional gesture analysis, affective computing, and biometrics. Her research has been reported by Finnish TV programs, newspapers and MIT Technology Review.
\vspace{-35mm}
\end{IEEEbiography}







\end{document}